\documentclass[11pt]{article}

\usepackage[T1]{fontenc}
\usepackage{mathptmx}

\usepackage[a4paper,margin=1in]{geometry}

\usepackage{amsmath,amssymb,amsthm,bbm}

\usepackage{graphicx}
\usepackage{tikz}
\usepackage{forest}
\usetikzlibrary{trees,positioning,shapes,shadows,arrows.meta}

\usepackage{booktabs,array,multirow,colortbl}

\usepackage{subcaption}
\usepackage{float}

\usepackage{xcolor,xspace,pifont}

\newcommand{\xmark}{\textcolor{red}{\ding{55}}}
\newcommand{\cmark}{\textcolor{green}{\ding{51}}}

\usepackage[font=small,labelfont=bf]{caption}

\usepackage[numbers,sort&compress]{natbib}

\usepackage[colorlinks,
            citecolor=blue,
            urlcolor=blue,
            linkcolor=blue]{hyperref}

\usepackage{authblk}

\theoremstyle{plain}
\newtheorem{thm}{Theorem}

\theoremstyle{definition}
\newtheorem{defn}[thm]{Definition}

\newcommand{\mstd}[2]{\ensuremath{#1{\scriptstyle \pm #2}}}

\newcommand{\MTTReNN}{\textsc{MT-TReNN}\xspace}
\newcommand{\TReNN}{\textsc{TReNN}\xspace}
\newcommand{\ReNN}{\textsc{ReNN}\xspace}
\newcommand{\TNN}{\textsc{TNN}\xspace}
\newcommand{\SNN}{\textsc{SNN}\xspace}

\newcommand{\col}{\cellcolor[HTML]{EFEFEF}}

\title{Boosting Team Modeling through Tempo-Relational Representation Learning}

\author[1]{Vincenzo Marco De Luca}
\author[1]{Giovanna Varni}
\author[1]{Andrea Passerini}

\affil[1]{Department of Information Engineering and Computer Science, University of Trento, Italy}

\affil[ ]{\texttt{
vincenzomarco.deluca@unitn.it,
giovanna.varni@unitn.it,
andrea.passerini@unitn.it}}

\date{} 

\begin{document}

\maketitle

\begin{abstract}
Team modeling remains a fundamental challenge at the intersection of Artificial Intelligence and Social Sciences.
Although a variety of computational models have been proposed in the last two decades, most fail to integrate Social Sciences insights, such as the critical role of temporal interactions in shaping team dynamics, and do not meet key practical requirements for real-world applications, including the ability to provide real-time, actionable recommendations to enhance team performance.
To address these limitations, in this paper, we propose a novel tempo-relational neural architecture that jointly models interactions between team members and the evolution of team dynamics through temporal graphs. We additionally propose a multi-task extension of the architecture that learns shared {\em social embeddings} for team members enabling the simultaneous prediction of multiple team constructs (e.g., Emergent Leadership, Leadership Style, and Teamwork components).
Experiments on two state-of-the-art team datasets show that our tempo-relational architecture outperforms temporal-only and relational-only approaches for team performance prediction, and that its multi-task extension substantially reduces training and inference time without loss of predictive performance. Finally, the integration of explainability techniques within the proposed architectures provides interpretable insights and actionable recommendations to support team improvement.
These strengths make our approach particularly well suited for human-centered artificial intelligence applications, such as intelligent decision-support systems in high-stakes collaborative environments.
\end{abstract}

\section{Introduction}
```

\begin{figure*}
\centering
\includegraphics[width=.9\textwidth]{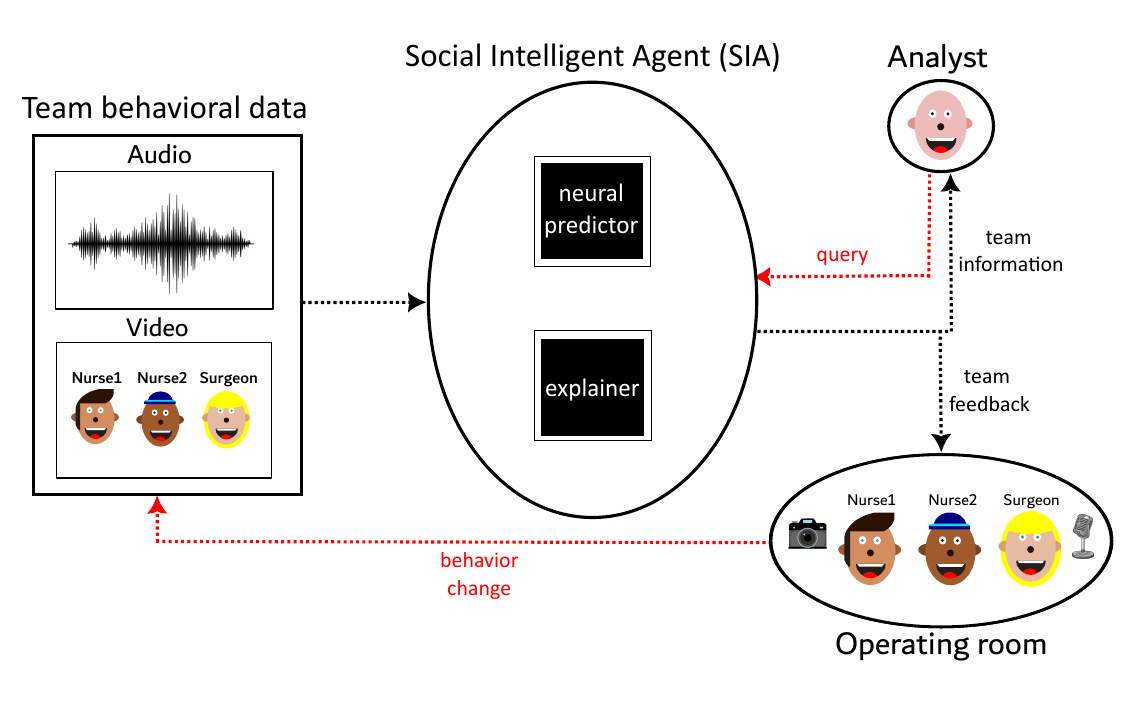}
\caption{
The SIA takes as input the team’s behavioral data (i.e., audio and video) directly captured in the operating room through sensors such as microphones and cameras. The SIA is composed of two modules: a neural predictor, used to infer team-related information, and an explainer, which generates feedback for the team. These outputs can be provided either offline or on the fly, either to an analyst or directly to the team in the operating room.
The red arrows indicate either the query performed by the analyst or the behavioral changes occurring in the scene and captured by the sensing devices.} 

\label{fig:complete_pipeline}
\end{figure*}

The increasing integration of Artificial Intelligence (AI) systems into everyday life has accelerated the shift from traditional human-computer interaction (HCI), where a single user acts as the decision maker, towards human-machine teaming (HMT) approaches~\cite{chignell2023evolution}. In HMT, groups of humans and AI agents collaborate, leveraging their complementary strengths to achieve shared goals~\cite{kaasinen2022smooth}. This collaboration has amplified the need to design and develop Human-Centered AI (HCAI) systems, characterized by transparency, trustworthiness, and collaborative intelligence~\cite{ozmen2023six}.
More specifically, among such systems, Socially Intelligent Agents (SIA)~\cite{wang2023steps} are specific agents equipped with social intelligence, namely the ability to navigate and succeed in social interactions by effectively engaging with others~\cite{Goleman2006-pj}.
This capability aligns with the broader vision of theory of mind AI~\cite{williams2022supporting}, which seeks to enable machines to infer, model, and reason about the mental states of others~\cite{cuzzolin2020knowing}.
To achieve such a goal, accurate and principled computational models of team interactions are needed~\cite{teaming2022state,westby2023collective, tavella2024towards}.
Although teams have been extensively studied within the Social Sciences, with particular focus on how members interact over time to give rise to emergent team processes (see, e.g.,~\cite{mcgrath1964social,steiner1972group,marks2001temporally, kozlowski-2006, mathieu2008team, kozlowski2018unpacking, klonek2024s}), key insights about team dynamics still remain largely unincorporated in computational models. Indeed, most computational models developed in Social Signal Processing~\cite{vinciarelli-2009, cristani2013human,palaghias2016survey, stergiou2019analyzing}, Affective Computing~\cite{chanel2015connecting, kaluarachchi2021review, de2022designing}, Reinforcement Learning~\cite{abbasi2022can}, or Large Language models~\cite{zu2024language, li2024socialgpt} still overlook social science findings. Thus, effective computational team modeling still remains an open challenge~\cite{Allen_2017, Reiter_2017, jackson2024modeling}.
Only a few computational approaches, indeed, capture the temporal evolution of team members' behavior, while many assume independence of observations across time, thereby neglecting dynamic interdependencies. Moreover, team members' relations are often oversimplified or ignored at the architectural level~\cite{maman2024modeling}, as they are reduced to concatenated or aggregated features~\cite{corbellini2022exploratory}. Another critical issue is that distinct yet interdependent constructs such as teamwork and leadership tend to be modeled separately, resulting in redundant learning pipelines~\cite{hung-gatica-perez-2010, murray2018predicting}, although recent works have begun to explore joint modeling approaches~\cite{sabry2021}. 
Finally, explainability remains limited: current models rarely provide transparent or actionable justifications for their predictions, despite growing demands for interpretability and accountability in HCAI~\cite{wang2019designing,goodman2017european}. 
Thus, there is a mismatch between established theoretical principles and their operationalization in computational models.  
This paper is a first step in bringing insights from Social Sciences on temporal dynamics, team members' relations, construct interdependence, and explainability, into a single unified computational approach. To this end, we propose two architectures: the \textit{Tempo-Relational Neural Network} (\TReNN), and its multi-task extension, the \textit{Multi-Task Tempo-Relational Neural Network} (\MTTReNN).  
From a methodological standpoint, these architectural choices not only close the main gaps identified in existing computational team modeling but also yield measurable scientific and practical advances.
The main contributions of this work are as follows:
\begin{itemize}
    \item A Tempo-Relational team computational approach (\TReNN): an architecture inspired by Social Sciences insights that models both team dynamics as well as team members' relations.
    It outperforms alternative strategies in predicting several team constructs (e.g., Teamwork components and Emergent Leadership) in multiple datasets;
    \item A multi-task extension of Tempo-Relational approach (\MTTReNN): an architecture that extends \TReNN to jointly predict multiple team constructs,     by learning robust \emph{social embeddings}, i.e., an embedding produced by a shared encoder in a multi-task setting. Such an embedding is designed to capture enough social information to generalize effectively across different team constructs. \MTTReNN offers the benefits of significantly reducing training and inference time, minimizing model storage requirements, while still producing accurate predictions;
    \item Two explainability modules: two modules that are seamlessly integrated in the architectures. 
    One module generates detailed, interpretable insights in the form of factual explanations, while the other provides actionable suggestions through counterfactual explanations. The validity of such insights and suggestions is supported by evidence drawn from data.    
\end{itemize}

These advances address key limitations of existing computational approaches. As a result, TRENN and MT-TRENN are strong candidates for HCAI applications in high-stakes collaborative environments. In such settings, a SIA is expected to act as a trustworthy teammate, supporting critical decision-making processes.
A compelling setting that illustrates the need for such a unified computational approach is the operating room (see Fig.~\ref{fig:complete_pipeline}), in which teamwork~\cite{catchpole2008teamwork, pronovost2013teamwork} and leadership~\cite{hu2016surgeons, parker2012surgeons} may critically affect surgical outcomes. During surgical procedures teams may commit errors due to lapses in coordination, miscommunication, or ineffective leadership. 
A SIA integrating our architectures can continuously monitor team behavior by analyzing multimodal input, such as paralinguistics, movement, and gaze. The agent predicts multiple team constructs both on-the-fly and offline. It also provides interpretable feedback on the drivers of suboptimal behaviors and supports informed debriefing procedures.
Such decision-support capabilities are essential in healthcare, where timely insights can help prevent errors and enhance patient safety. Although the operating theatre serves as a vivid example of high-stakes, error-prone environments~\cite{brown2001err, makary2016medical}, our unified computational framework is broadly applicable to any human-machine teaming system.

The remainder of this paper is organized as follows: Sec.~\ref{sec:rel_works} reviews related work ranging from static team modeling approaches to temporal, relational or multi-task approaches; Sec.~\ref{background} reports foundational team-related definitions in Social Sciences. Next, Sec.~\ref{sec:snn_to_trenn} describes the evolutionary trajectory of neural paradigms for team modeling, culminating in our novel \TReNN and \MTTReNN architectures, which are discussed in further detail in Sec.~\ref{sec:method}. Then, Sec.~\ref{expres} reports experimental evaluation of \TReNN, \MTTReNN, and the generation of factual and counterfactual explanations. Finally, Sec.~\ref{sec:conclusion} draws conclusions and outlines limitations and future work.

\section{Related Work}
\label{sec:rel_works}

Automated team modeling has gained increasing attention over the past two decades, driven by the need to develop systems capable of understanding, modeling, and automatically interpreting human social interactions.
Progress in this field, however, has been constrained by the technical difficulties of collecting multimodal data in multiparty scenarios as well as by the scarcity of publicly available datasets (e.g.,~\cite{kraaij2005ami, sanchez2011audio, beyan2016detecting, braley2018group, bhattacharya2019unobtrusive, muller2021multimediate}).
In this section, we review previous work on computational team modeling according to the aspects chosen by the authors to model team behavior:
(i) static team modeling; 
(ii) temporal team modeling; 
(iii) relational team modeling; and 
(iv) multi-task and multi-construct team modeling.
\subsection{Static team modeling}
Static team models rely on individual-level or aggregated multimodal features (e.g., speech, video) to model team-level constructs without explicitly modeling temporal evolution or interactions among team members.
Within this paradigm, some of these computational approaches extract behavioral features from one or two modalities and then pool them across team members. The pooled features, computed as a simple 
average or concatenation~\cite{enayet2021analyzing, yang2023multimediate}, form a team-based feature vector that serves as input to machine learning models.
The primary objective of these studies is to investigate which features are most informative to predict a specific team construct. 
One of the earliest works in static team modeling was proposed by Hung and colleagues in 2010~\cite{hung-gatica-perez-2010}. They fed hand-crafted audio and visual features into a naive Bayes classifier and a support vector machine to estimate team cohesion.
Later studies have also predicted team cohesion, introducing pairwise and team-level features derived from the alignment of para-linguistic speech behavior and fed into a Gaussian mixture model and a kernel density estimator~\cite{nanninga-2017}.
Related works have explored leadership emergence and its connection to the team members' body pose patterns using support vector machines~\cite{beyan2017moving}.
In a similar static fashion, acoustic features and linguistic features extracted from conversation transcripts have been used as input for gradient boosting and extra-trees~\cite{murray2018predicting, kubasova2019analyzing} to predict team performance in task-based interactions. 
Moreover, engagement has been modeled through multimodal features (i.e., speech and visual features) via feed-forward neural networks (FFNN)~\cite{muller2024multimediate}.
Despite the effort of interdisciplinary research, most computational models
still do not encode time and relations in their analysis, as shown in recent surveys (e.g.,~\cite{rasipuram2020automatic}).

\textit{\TReNN addresses these limitations by explicitly modeling temporal dynamics in the predictive architecture.}

\subsection{Temporal team modeling}
Temporal team models incorporate the evolution of individual behavior over time but treat team members independently, without modeling interpersonal relations or cross-team construct dependencies.
Social sciences have long demonstrated that team communication patterns can be described as temporal graphs~\cite{bavelas1950communication}. However, only a few studies explicitly integrate the temporal information into the computational modeling~\cite{lin2020predicting}. 
Early work introducing temporal information proposed an unsupervised approach to model parallel episodes~\cite{10.1145/3279981.3279983} that captures temporally proximal events and overlapping actions, but encoding time only at the individual feature level.
An alternative line of research draws from dynamical systems theory~\cite{mcgrath1997small, gorman2017understanding}, which provides a theoretical foundation to understand team behavior as evolving patterns over time.
Building on this foundation, more recent works adopt multimodal data-driven sequence-to-sequence modeling strategies~\cite{sutskever2014sequence}, including Hidden Markov Models~\cite{avci2016predicting},  Conditional Restricted Boltzmann Machines~\cite{8626131}, Long-Short Term Memory (LSTM)~\cite{lin2023interaction}, and Fusion Transformers~\cite{10191640}.
To further leverage hierarchical temporal patterns, recurrent architectures such as HPNet~\cite{qiu2019neurally} have been proposed for modeling multi-scale sequences in spatio-temporal data like video frames. 
Other approaches model discrete symbolic actions through Bayesian approaches to predict team mental states~\cite{seo2021towards}, or leverage Multi-Agent Imitation Learning to learn a generative model of team behavior~\cite{seo2023automated} validated for intervention to enhance team performance~\cite{seo2025socratic}.
Similarly, architectures such as Spatial-TeamFormer~\cite{yu2022learning} and GroupFormer~\cite{li2021groupformer} model temporal patterns in multi-agent settings, relying on visual, spatial, or trajectory data. 
Despite these advances, most team constructs influenced by interactions among members are still modeled without capturing the relational structure~\cite{junior2019first}.

\textit{\TReNN addresses the limitations of temporal models by jointly modelling temporal and relational information.}

\subsection{Relational team modeling}
Relational team models incorporate information about how team members influence one another, but ignore the team dynamics over time and the interdependencies among team constructs.
Early proposals integrate relations only at the feature level by combining individual-level and team-level features, but these approaches apply Convolutional or Recurrent Neural Networks independently to each team member.
Scores as team-member coefficients in the cost function have also been proposed during the optimization process applied to Random Forest outcomes.~\cite{10.1145/3382507.3418877}. 
Attempts to integrate relations at the modeling level were based on network analysis~\cite{leenders2016once, chai2019applying}, and complex systems~\cite{dooley1997complex, arrow2000small, ramos2018teams}. 
The first work~\cite{lin2020predicting} adopting Graph Neural Networks (GNNs)~\cite{gori2005new, scarselli2008graph} uses acoustic and verbal as individual features and speaking turns as edges modeled through Graph Convolutional Networks (GCNs)~\cite{kipf2016semi} to predict only team performances.
Sharma and colleagues~\cite{sharma2023graphitti} proposed hierarchical GNNs, fed with visual features, to predict the dominance score and ranking of team members. 
In recent years, Temporal GNNs (TGNNs) have been proposed to solve tasks such as 
behavior forecasting~\cite{yang2020group}, impression prediction~\cite{bai2023dips}, and human interaction classification~\cite{malik2023relational}.
A recent work proposes Cross-Attention~\cite{tan2019lxmert} between sender and receiver to model interaction in engagement prediction~\cite{li2024dat}.
Moreover, time modeling through Imitation Learning and hierarchical structure has been proposed to predict individual behavior and team behavior~\cite{seo2025hierarchical}, but still lacks an explicit modeling of relations. Hypergraphs for recognizing social interaction~\cite{tang2025hypergraph} and GNNs combined with Multimodal Large Language Models~\cite{tang2025cross} are other additional works that leverage GNNs for teams, but without assessing the quality and behavior of the team.
Some recent probabilistic frameworks take a different approach to modeling team relations. 
For example, Bayesian Team Cognition~\cite{westby2023collective} infers latent cognitive states of team members from observed decisions and communications. 
While this approach provides interpretable insights into teammate beliefs and intentions, it does not produce continuous multi-task behavioral constructs.
To the best of our knowledge, no work addresses team process prediction using TGNN architectures yet.

\textit{As for the purely temporal approaches, \TReNN overcomes the limitations of purely relational approaches thanks to its joint tempo-relational modeling capabilities.}
\subsection{Multi-task team modeling}
Multi-task team models aim to predict multiple team constructs from individual team member features.
They leverage dependencies among constructs by projecting member features into a shared embedding space.
This space is later specialized through task-specific fully connected networks. Despite this advance in modeling multiple team constructs, they still disregard the temporal evolution of the team and the importance of modeling relations in the team. 
Approaches mentioned so far develop separate models to predict different team constructs or different dimensions of the same construct.
Consequently, each dimension requires training a model from scratch.
This design results in increased spatial and temporal complexity.
The number of parameters grows with the number of tasks, and training multiple independent models increases training time.
Moreover, these approaches fail to address real-world application requirements.
In practice, agents must jointly reason over multiple interlaced team constructs to effectively collaborate with human teams~\cite{sebo2020robots, chugunova2022we, dahiya2023survey}. 
Beyan et al.~\cite{beyan2017multi} introduced multi-task learning (MTL) to predict leadership styles in team interaction. Such an approach, however, completely disregards dynamics and relations among team members. Ghosh et al. ~\cite{ghosh2020automatic} propose an approach to predict cohesion through MTL, whereas Maman et al. ~\cite{maman2020game} present two architectures   
to jointly predict social and task cohesion and the valence of emotion. 

Other works explore the possibility of generating social embeddings that can be reused for multiple tasks, but restrict this reuse to fine-tuning strategies~\cite{morgan2021classifying}.
As a result, additional parameters are introduced and task-specific specialization remains necessary for doing inference on other tasks.

\textit{\MTTReNN improves over purely static multi-task team modeling approaches by incorporating multi-task learning functionalities in the tempo-relational architecture developed with \TReNN}.

\begin{table*}[t]
\centering
\renewcommand{\arraystretch}{1.4}
\begin{tabular}{
  >{\raggedright\arraybackslash}m{0.30\textwidth}
  | >{\raggedright\arraybackslash}m{0.65\textwidth}
}
\hline
\textbf{Term} & \textbf{Definition} \\
\hline
\textit{Team} &
A \textit{team} is a social unit of more than two individuals with specified roles who interact adaptively, interdependently, and dynamically toward a common and valued goal~\cite{dyer1984team,salas1992toward,moreland2010}. \\
\hline
\textit{Team construct} &
A \textit{team construct} is any measurable property capturing a specific behavioral, cognitive, affective, or relational aspect of team functioning (e.g., teamwork, leadership style). \\
\hline
\textit{Team interaction} &
\textit{Team interaction} refers to the exchange of information, behaviors, and affect among team members as they coordinate, communicate, and collaborate to achieve shared goals~\cite{cooke2013interactive}. \\
\hline
\textit{Team dynamics} &
\textit{Team dynamics} concerns the evolution of team behaviors and member interactions over time, leading to the emergence of complex collective social phenomena~\cite{delice2019advancing}. \\
\hline
\end{tabular}
\caption{Definitions of key team-related concepts relevant to computational team modeling.}
\label{tab:team_definitions}
\end{table*}

\begin{table*}[t]
\centering
\renewcommand{\arraystretch}{1.5}
\begin{tabular}{
  >{\raggedright\arraybackslash}m{0.30\textwidth}
  | >{\raggedright\arraybackslash}m{0.65\textwidth}
}
\hline
\textbf{Main Component} & \textbf{Definition} \\
\hline
\textit{Adaptability (A)} &
The ability of a team to adjust roles, strategies, and resource allocation in response to environmental changes~\cite{salas2005there}. \\
\hline
\textit{Back-up behavior (BB)} &
The ability of team members to anticipate the needs of others and provide timely support when required~\cite{salas2005there}. \\
\hline
\textit{Mutual performance monitoring (MPM)} &
The ability to comprehend the team environment and continuously monitor team members’ performance to ensure effective coordination~\cite{salas2005there}. \\
\hline
\textit{Team leadership (TL)} &
The capacity to direct, coordinate, and guide the activities of team members toward shared objectives~\cite{salas2005there}. \\
\hline
\textit{Team orientation (TO)} &
The capacity of team members to prioritize collective goals and outcomes over individual interests~\cite{salas2005there}. \\
\hline
\end{tabular}
\caption{Core behavioral components of the Big Five in Teamwork framework.}
\label{tab:main_tw_components}
\end{table*}
\begin{table*}[t]
\centering
\renewcommand{\arraystretch}{1.5}
\begin{tabular}{
  >{\raggedright\arraybackslash}m{0.30\textwidth}
  | >{\raggedright\arraybackslash}m{0.65\textwidth}
}
\hline
\textbf{Coordinating Mechanism} & \textbf{Definition} \\
\hline
\textit{Closed-loop communication (CC)} &
The process by which team members explicitly direct, acknowledge, and confirm messages to ensure accurate receipt and shared understanding~\cite{salas2005there}. \\
\hline
\textit{Mutual trust (MT)} &
The shared belief among team members that others will reliably fulfill their roles and act in the best interest of the team~\cite{salas2005there}. \\
\hline
\textit{Shared mental models (SMM)} &
The common and overlapping understanding among team members regarding tasks, roles, and interaction patterns that supports effective coordination and anticipation of teammates’ actions~\cite{salas2005there}. \\
\hline
\end{tabular}
\caption{Coordinating mechanisms supporting the main teamwork components according to the Big Five in Teamwork framework.}
\label{tab:coordinating_mechanisms}
\end{table*}

\section{Background}
\label{background}

\begin{figure}[!h]
\centering
\includegraphics[width=0.45\textwidth]{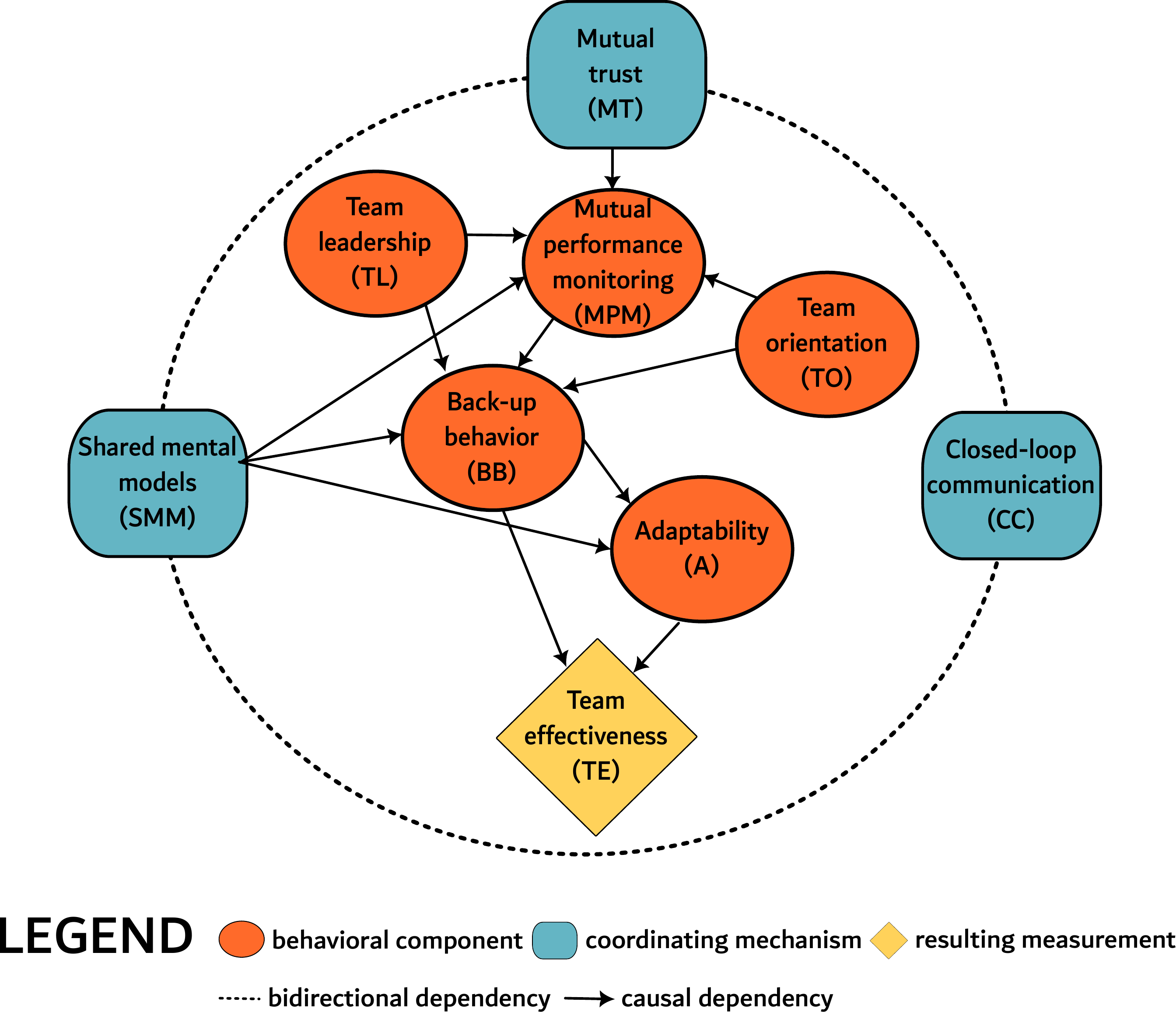}
\caption{The BFT framework. It includes five core components (in orange), 
and three additional components (in blue), representing coordinating mechanisms, 
All components collectively contribute to Team Effectiveness (TE), which is represented in the yellow node. Arrows indicate direct influences between connected components.}
\label{fig:BIG_FIVE_TEAMWORK}
\end{figure}

From a Social Sciences perspective, team effectiveness emerges from the interplay of relational structures and situational forces.
Relational Models Theory ~\cite{fiske1991structures, fiske1992relativity} offers a pathway to understand these relational structures. It describes how social behavior and the quality of the interactions are guided by shared relational expectations, such as cooperation among peers, hierarchical coordination, or reciprocity in exchanges. These expectations, in turn, shape collaboration, trust, and the emergence of leadership.
Field Theory~\cite{lewin1948field} complements this view by conceptualizing team behavior as a function of individual characteristics operating within a dynamic social field shaped by group norms, shared goals, and  patterns of interpersonal influence. Together, these perspectives ground teamwork processes in both relational structures and situational context. 
To systematically analyze the team framework, we first introduce the fundamental concepts that underpin team theory. Table~\ref{tab:team_definitions} presents clear definitions of the fundamental team-related concepts serving as a reference point for the frameworks and mechanisms discussed in this section.
Building on this foundation, the Big Five in Teamwork (BFT) framework~\cite{salas2005there} identifies five behavioral components, defined in Table~\ref{tab:main_tw_components}, supported by three coordinating mechanisms, listed in Table~\ref{tab:coordinating_mechanisms}, which collectively map individual and collective behaviors into a unified measure of Team Effectiveness (TE). 
\begin{defn}
\textit{Teamwork (TW) is the set of tied thoughts, actions, and feelings of each team member that are needed to function as a team and that combine to facilitate coordinated, adaptive performance and task objectives resulting in value-added outcomes}~\cite{morgan1986measurement, Salas2004CooperationAW, salas2005there}.
\end{defn}
BFT conceptualizes TW as a set of interrelated mechanisms, as illustrated in Fig.~\ref{fig:BIG_FIVE_TEAMWORK}: Mutual Trust (MT) is affected by the communication among team members (CC), and influences the presence of Shared Mental Models (SMM) among team members. The Mutual Performance Monitoring (MPM) is guaranteed by the Team Leadership (TL), the Orientation of the Team (TO), and MT. Proper Back-up Behavior (BB) requires effective TL, SMM, MPM, and TO, while Adaptability (A) requires proper BB and SMM. 
BFT presents ten propositions that define how these team components contribute to TE.
In particular, TL influences TE by setting performance expectations, including performance monitoring and backup behavior.
MPM affects TE through effective BB and requires adequate SMM and MT among team members.
BB directly contributes to TE by ensuring task completion, although its effects are mediated by the team’s ability to adapt to internal and external changes.
Effective BB further requires adequate SMM and MPM.
A directly affects TE and depends on SMM, MPM, and BB.
Finally, TO influences TE through members’ willingness to mutually monitor performance and accept support from others through BB.
These interdependent mechanisms illustrate how coordination, trust, and monitoring naturally give rise to patterns of influence within a team, implying that certain team members influence team processes more strongly due to their behavior, relational positioning, and interactions. This perspective provides a smooth conceptual bridge from the competencies captured by BFT to the study of leadership dynamics.

\begin{defn}
\label{def:glis}
 \textit{Emergent Leadership} (EL) is an emergent state that characterizes a team member who naturally appears as the leader during social interaction, without any formal leadership role, and whose power comes from peers in the team, rather than from an established authority~\cite{stein1975identifying}.
\end{defn}

Given the pivotal role of leadership, Field Theory and Relational Models Theory jointly suggest that different Leadership Styles can foster or hinder key team constructs including communication, coordination, trust-building,  adaptability, and EL~\cite{ford2008perceived}. These effects ultimately impact overall team performance and decision-making~\cite{zaccaro2001team, burke2006type, sauer2011taking, dinh2014leadership}. 

\begin{defn}
\label{def:leadership_style}
The \textit{Leadership Style} (LS) reflects how a team member, i.e., the leader, interacts with others when exhibiting leadership~\cite{bales1980symlog}.
\end{defn}

The interplay of TW, EL, and LS encapsulates fundamental aspects of team functioning, shaping how teams coordinate, adapt, and perform over time~\cite{kozlowski1996dynamic, kozlowski2006enhancing, derue2010will, i2014leader}.
Developing computational team models that predict these team constructs over time, and that analyze how individual member and interactions influence them, provides a valuable asset for both advancing Social Sciences theories and designing innovative human-machine teaming applications.

\section{From Static to Multi-task Tempo Relational Team Modelling}
\label{sec:snn_to_trenn}

This section first outlines the evolutionary trajectory of neural paradigms for team modeling.
We first formalize the existing approaches into a taxonomy of static, temporal, and relational models (Sec.~\ref{sub:existing_approaches}).
We then present our proposed \TReNN approach that jointly captures temporal and relational information, and its extension \MTTReNN designed for the simultaneous prediction of multiple interleaved team constructs through a MTL objective (Sec.~\ref{sub:proposed_approaches}). Implementation details of \TReNN and \MTTReNN are reported in Sec.~\ref{sec:method}.

We formalize all architectures within a encoder–decoder abstraction for team modeling.
The encoder produces a social embedding of team members, which is then decoded into predictions of team constructs.
The encoder $f(\cdot)$ maps an input vector $\textbf{x} \in \mathbb{R}^{n}$, representing $n$ individual behavioral features
per team member, into a hidden representation $\textbf{h} \in \mathbb{R}^{d}$.
The decoder $g(\cdot)$ maps this hidden representation $\textbf{h} \in \mathbb{R}^{d} $ into an output vector $\mathbf{y} \in  \mathbb{R}^{m}$, representing $m \geq 1$ team construct scores.

\begin{defn}\label{def:social_embedding} The \emph{social embedding}  $\textbf{h} \in \mathbb{R}^{d}$ is the
  hidden representation generated by the encoder $f(\cdot)$. It is designed to capture individual and relational information from input data which is informative enough for the decoder $g(\cdot)$ to predict one or more team constructs $\textbf{y}$ for each team member.
\end{defn}

Table~\ref{tab:notation} summarizes the mathematical notation used throughout this section.

\begin{table}[t]\label{tab:notation}
\centering
\small
\begin{tabular}{ll}
\toprule
\textbf{Symbol} & \textbf{Description} \\
\midrule
\multicolumn{2}{l}{\emph{Indices and dimensions}} \\
\midrule
$v,u,i$ & Team member (node) indices \\
$t$ & Time index \\
$K$ & Temporal window length \\
$t:t+K$ & Temporal interval from $t$ to $t+K$ \\
$k$ & Task index in multi-task learning \\
$n$ & Number of team members ($|\mathcal{V}|$) \\
$d$ & Feature or embedding dimensionality \\
$m$ & Number of team constructs (tasks) \\

\midrule
\multicolumn{2}{l}{\emph{Feature representations}} \\
\midrule
$\mathbf{x}$ & Feature vector of a single team member \\
$\mathbf{x}_v^t$ & Features of team member $v$ at time $t$ \\
$\mathbf{X}$ & Matrix of all team members' features \\
$\mathbf{X}^t$ & Feature matrix at time $t$ \\

\midrule
\multicolumn{2}{l}{\emph{Embeddings and outputs}} \\
\midrule
$\mathbf{h}_v$ & Static social embedding of team member $v$ \\
$\mathbf{h}_v^t$ & Social embedding of member $v$ at time $t$ \\
$\mathbf{h}_v^{t:t+K}$ & Temporal embedding over interval $[t,t+K]$ \\
$\mathbf{y}_v$ & Output prediction for team member $v$ \\
$\mathbf{y}_v^k$ & Prediction of task $k$ for member $v$ \\
$\mathbf{Y}$ & Matrix of outputs for all members and tasks \\

\midrule
\multicolumn{2}{l}{\emph{Graphs and neighborhoods}} \\
\midrule
$\mathcal{V}$ & Set of team members (nodes) \\
$\mathcal{E}$ & Set of interactions (edges) \\
$\mathcal{N}(v)$ & Neighborhood of node $v$ \\
$G^t=(\mathcal{V},\mathcal{E},\mathbf{X}^t)$ & Team graph at time $t$ \\
$\mathbf{G}=\{G^t,\dots,G^{t+K}\}$ & Dynamic team as temporal graph sequence \\

\midrule
\multicolumn{2}{l}{\emph{Functions}} \\
\midrule
$f(\cdot)$ & Encoder function \\
$g(\cdot)$ & Decoder function \\
$g_{\text{mt}}(\cdot)$ & Multi-task decoder \\

\bottomrule
\end{tabular}
\caption{Summary of the notation used for static, temporal, relational, and multi-task team modeling.}
\label{tab:notation}
\end{table}

\begin{figure*}[htbp]
    \begin{center}
    \begin{tabular}{c c c}
        \subfloat[\SNN]{\includegraphics[width=0.2\textwidth]{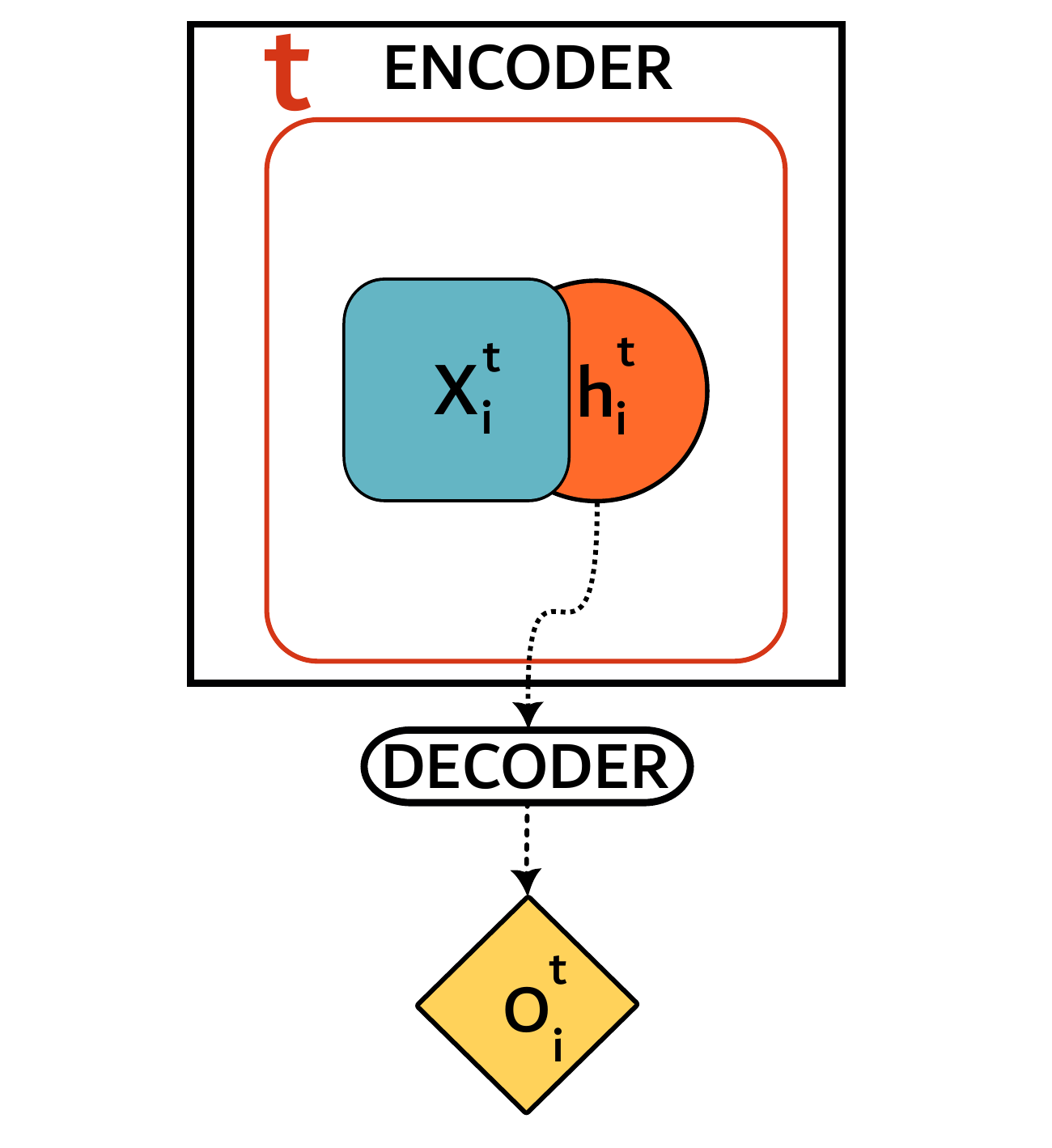} \label{fig:snn}} & 
        \subfloat[\TNN]{\includegraphics[width=0.4\textwidth]{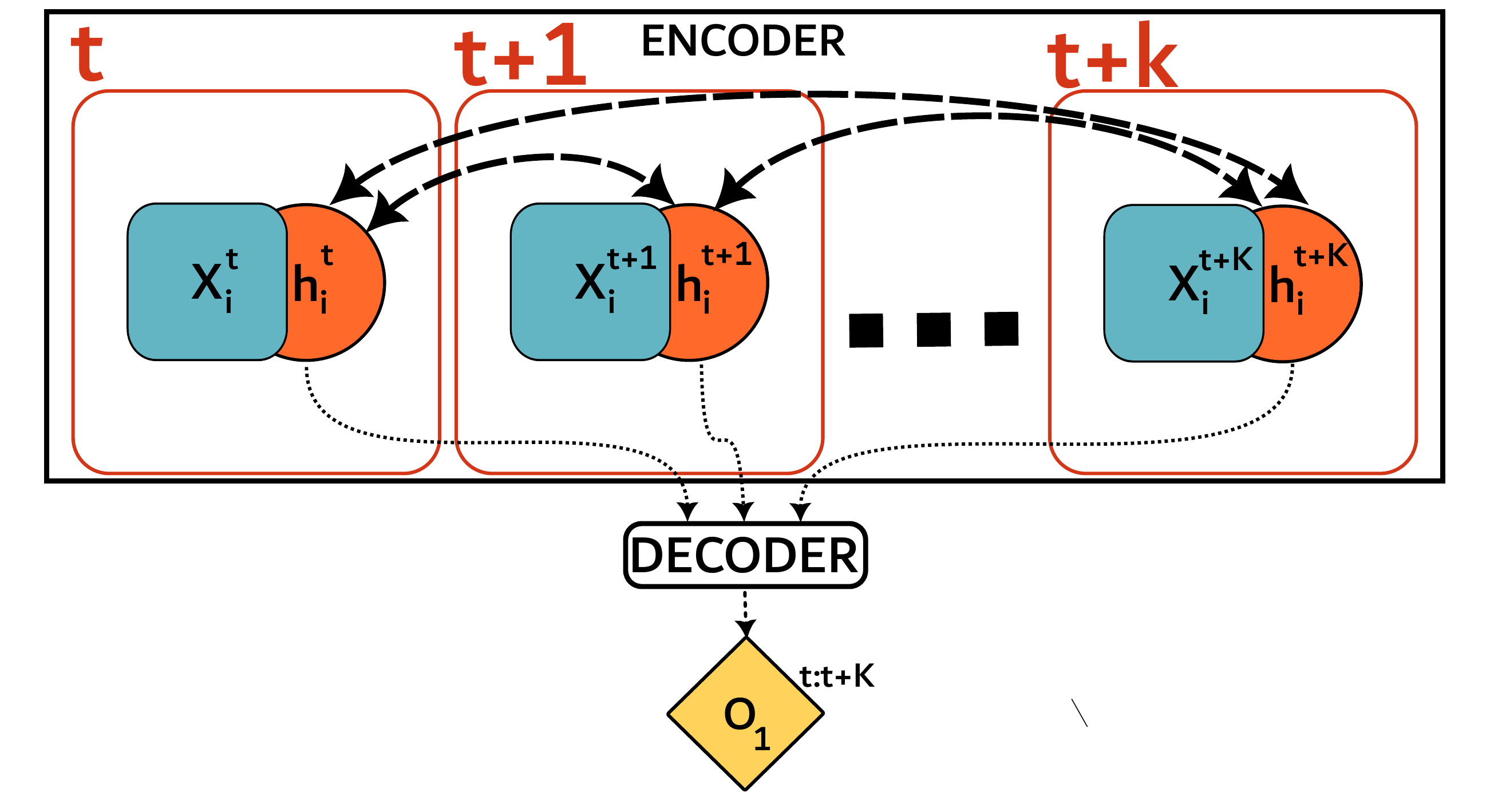} \label{fig:tnn}} & 
        \subfloat[\ReNN]{\includegraphics[width=0.2\textwidth]{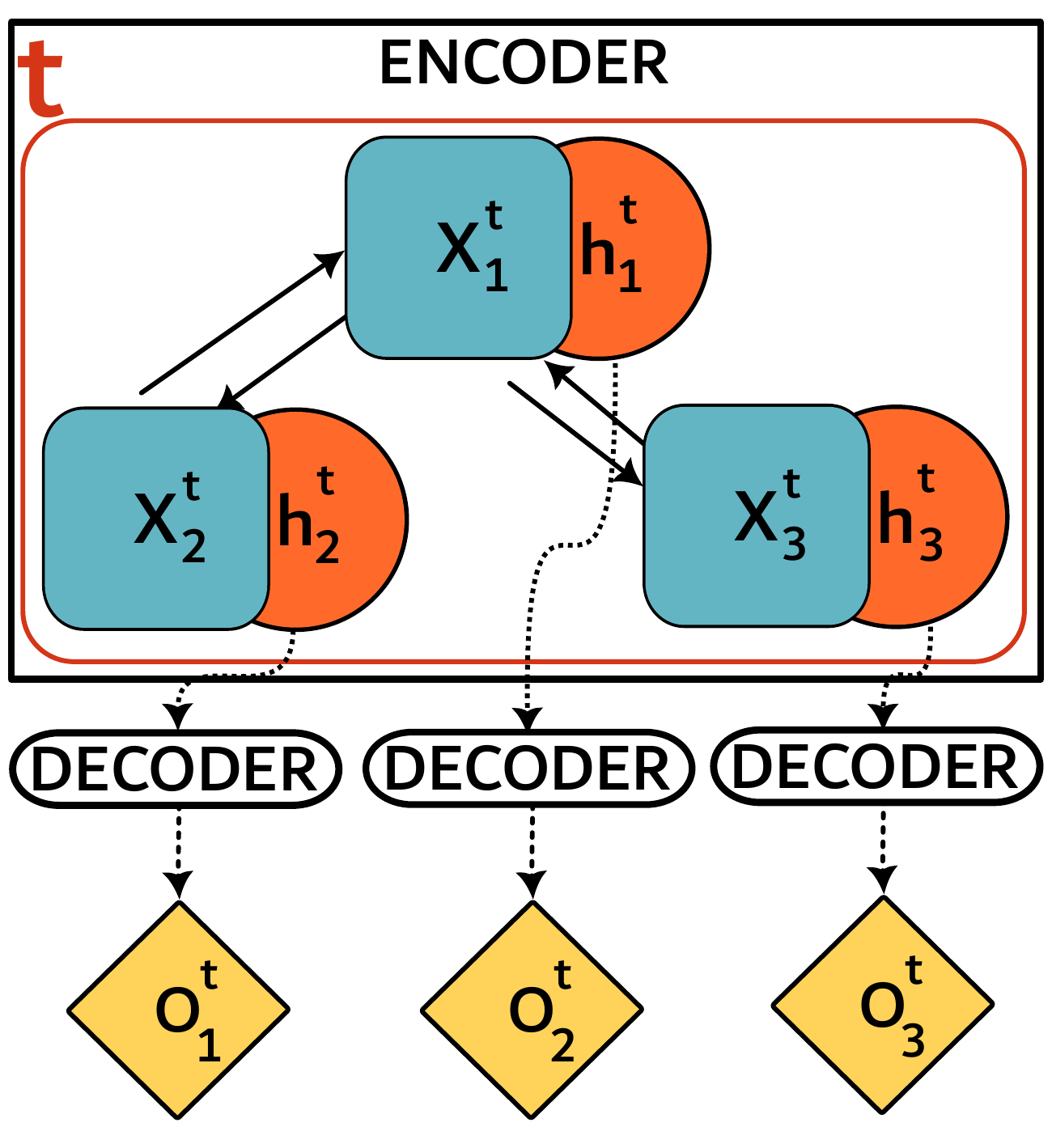} \label{fig:gnn}} \\
    \end{tabular}
    
    \vspace{1em}
    \centering
    \subfloat[\TReNN]{\includegraphics[width=0.8\textwidth]{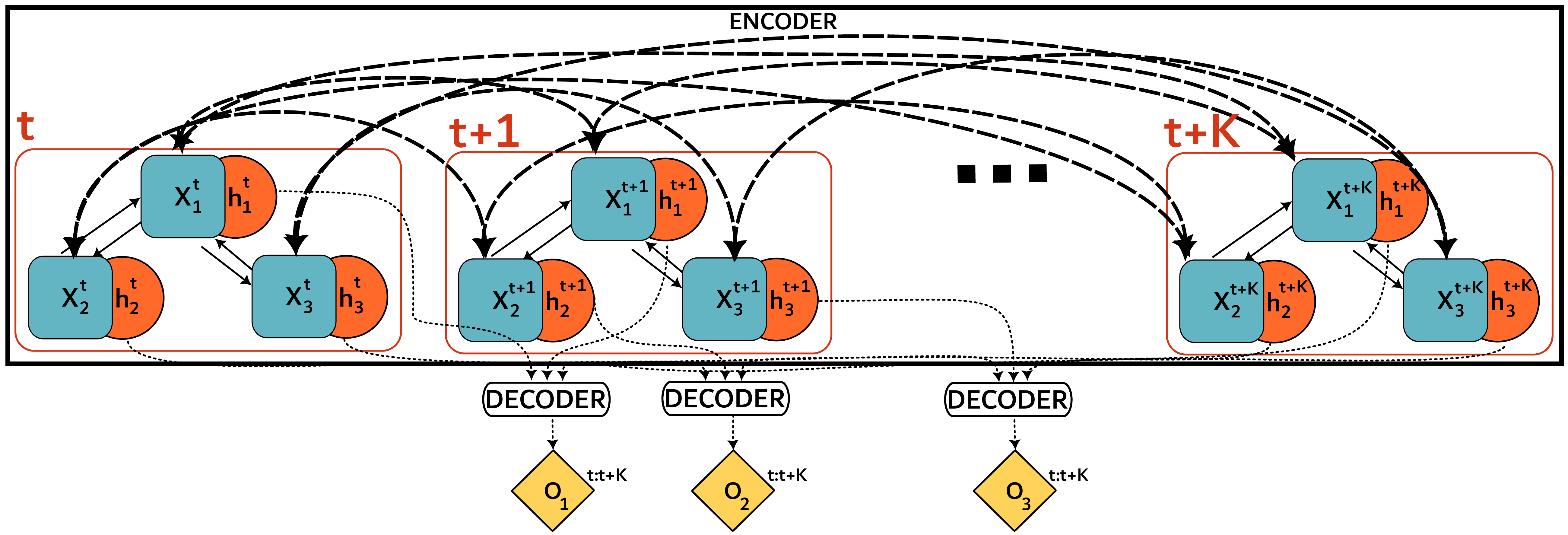} \label{fig:tgnn}}

    \vspace{1em}
    \centering
    \subfloat[\MTTReNN]{\includegraphics[width=0.8\textwidth]{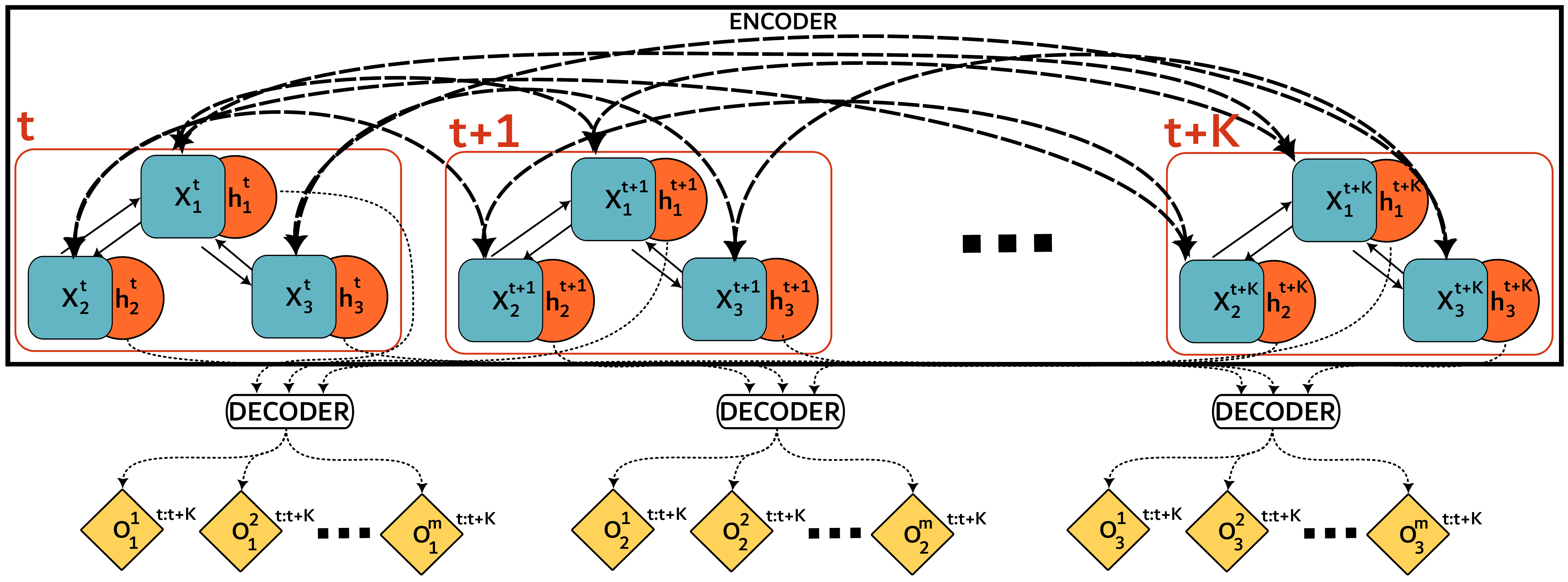} \label{fig:mttgnn}}
    
    \vspace{0.5em}
    \begin{minipage}{\textwidth}
        \raggedright
        \includegraphics[width=0.8\textwidth]{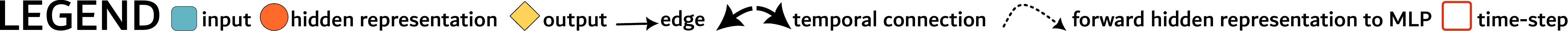}
        \label{fig:legend}
    \end{minipage}
    \end{center}
    \caption{Neural Network-based architectures to model teams. 
    Rectangular boxes with red borders are \emph{encoders}, while round-shaped boxes with black borders are \emph{decoders}. Any encoder contains the input data $X_{j}^{t+i}$, and its corresponding hidden representation $h_{j}^{t+i}$, where the subscript $j$ refers to the team member and the superscript $t+i$ refers to the timestep. Any decoder receives as input the hidden representation $h_{j}^{t+i}$, and maps it into the output space $o_{j}^{k,t+i}$, where the subscript $j$ refers to the team member and the superscript $k$ refers to the predicted team task, and $t+i$ the timestep. (a) \SNN takes as input team member information at a specific time-step $X_{i}^{t}$, and maps it into its corresponding hidden representation $h_{i}^t$. Then, the decoder from the hidden representation predicts the output only based on team member and instantaneous information; (b) \TNN: the encoder takes as input a sequence of team member information $(X_{i}^{t}, X_{i}^{t+1}, ..., X_{i}^{t+K})$; the temporal evolution of team member information is leveraged to produce the hidden representation $\left(h_{i}^{t}, h_{i}^{t+1},..., h_{i}^{t+K}\right)$. Next, the decoder from the hidden representation gets the final prediction $o_{i}^{t:t+K}$ that is based on the temporal evolution of the team member; (c) \ReNN: the encoder takes as input the team member information as well as the team member interactions at a specific time step. Let us assume a team composed of three people: $X_{1}^{t}, X_{2}^{t}, X_{3}^{t}$, the encoder produces the corresponding hidden representation for each of them: $h_{1}^{t}, h_{2}^{t}, h_{3}^{t}$, while the \emph{decoder} receives from such a hidden representation outputs the prediction corresponding to each team member $o_{1}^{t}, o_{2}^{t}, o_{3}^{t}$; (d) \TReNN: the encoder takes as input the team member information as well as the team members interactions over multiple time-steps. Let us assume a team composed of three people: $X_{1}^{t+i}, X_{2}^{t+i}, X_{3}^{t+i}, \forall i \in [0,K]$, the encorder produces the corresponding hidden representation for each of them: $h_{1}^{t+i}, h_{2}^{t+i}, h_{3}^{t+i}$, while the decoder from such a hidden representation gets the prediction corresponding to each team member $o_{1}^{t+i}, o_{2}^{t+i}, o_{3}^{t+i}$;(e) \MTTReNN: the encoder works exactly as in \TReNN, while the decoder learns multiple team-tasks: given the hidden representation, it generate, for each team member, the prediction of multiple team-tasks $o_{1}^{1,t+i}, o_{2}^{2,t+i}, ..., o_{3}^{m,t+i}$}.
    \label{fig:architecture_comparison}
\end{figure*}

\subsection{Static and temporal/relational modeling}
\label{sub:existing_approaches}
Existing approaches range from static, independent representations of team members to models that incorporate either temporal dynamics or interpersonal interactions.

Based on the assumption that individual team member behaviors suffice to predict team constructs, the simplest encoder is \emph{Static Neural Network} (\SNN)~\cite{hung-gatica-perez-2010,nanninga-2017, murray2018predicting, kubasova2019analyzing} implemented as FFNN, as shown in Fig.~\ref{fig:snn}, that models a \emph{static team member}.
\begin{defn}\label{def:static_team_member}
A \emph{static team member} is defined as a pair $\{\textbf{x}, \textbf{y}\}$, where $\textbf{x} \in \mathbb{R}^{d}$ are input features, and $\textbf{y} \in \mathbb{R}$ is the output score.
\end{defn}
\begin{defn}\label{def:snn}
A \SNN uses its \emph{encoder} $f_{\text{snn}}$ to map a \emph{static team member} into their corresponding embedding $\textbf{h}$ (Eq.~\ref{eq:snn}) and then uses its \emph{decoder} $g$ to map this embedding into the output space $y \in \mathbb{R}$ (Eq.~\ref{eq:decoder_traditional}).
\begin{equation}
\textbf{h} = f_{\text{snn}}(\textbf{x}) 
\label{eq:snn}
\end{equation}
\begin{equation}
\textbf{y} = g(\textbf{h})
\label{eq:decoder_traditional}
\end{equation}
\end{defn}
\SNN models each team member in isolation, ignoring temporal and relational context as well as multiple team constructs.
Consequently, SNN fails to capture the behavioral dynamics essential in team settings, resulting in an oversimplified representation of complex social interactions~\cite{yang2023multimediate}.
To overcome these limitations, researchers extended \SNN by modeling individual behavior over time and leveraging principles from dynamical systems~\cite{mcgrath1997small,gorman2017understanding} or adopting Temporal Neural Networks (\TNN) implemented as Recurrent Neural Networks, LSTM~\cite{schmidhuber1997long}, Gated-Recurrent Unit~\cite{cho2014gru}, 
or Multi-Head Attention (MHA)~\cite{vaswani2017attention}.

\newcolumntype{C}[1]{>{\centering\arraybackslash}m{#1}} 
\newcolumntype{L}[1]{>{\raggedright\arraybackslash}m{#1}} 

\begin{table*}[t]
    \centering
    \scalebox{0.99}{%
    \begin{tabular}{|C{4.6cm}|C{1.7cm}|C{1.7cm}|C{1.7cm}|C{2.3cm}|C{2.3cm}|}
    \toprule
    \multirow{2}{*}{\textbf{Architecture}} 
    & \multicolumn{3}{c|}{\textbf{Type of modeling}} 
    & \multicolumn{2}{c|}{\textbf{Input-Output mapping}} \\
    \cmidrule(r){2-4} \cmidrule(l){5-6}
    & \textbf{Relational} & \textbf{Temporal} & \textbf{Multi-task} & \textbf{Input} & \textbf{Output} \\
    \midrule
            \multicolumn{1}{|L{4.6cm}|}{\SNN \cite{hung-gatica-perez-2010, nanninga-2017,beyan2017moving,murray2018predicting,kubasova2019analyzing,rasipuram2020automatic,enayet2021analyzing, yang2023multimediate,muller2024multimediate}}      & \xmark & \xmark & \xmark & \multicolumn{1}{|L{2.3cm}|}{$x \in \mathbb{R}^{d}$} & \multicolumn{1}{|L{2.3cm}|}{$y \in \mathbb{R}$} \\
\multicolumn{1}{|L{4.6cm}|}{\TNN \cite{avci2016predicting, gorman2017understanding, 8626131, junior2019first, lin2020predicting, seo2021towards, 10191640, lin2023interaction, seo2023automated, seo2025socratic}}     & \xmark 
& \cmark
& \xmark
& \multicolumn{1}{|L{2.3cm}|}{$x \in \mathbb{R}^{t \times d}$} 
& \multicolumn{1}{|L{2.3cm}|}{$y \in \mathbb{R}$} \\
\multicolumn{1}{|L{4.6cm}|}
{\ReNN \cite{10.1145/3338245,10.1145/3382507.3418877,
lin2020predicting,chien2021self,chien2021belongingness,sharma2023graphitti,tang2025hypergraph,tang2025cross}} 
& \cmark 
& \xmark 
& \xmark 
& \multicolumn{1}{|L{2.2cm}|}{$G=(\mathbb{V}, \mathbb{E}, X)$}
& \multicolumn{1}{|L{2.2cm}|}{$y \in \mathbb{R}^{|\mathbb{V}|}$} 
\\
\multicolumn{1}{|L{4.4cm}|}
{\textbf{\TReNN}}       
& \cmark 
& \cmark 
& \xmark
& \multicolumn{1}{|L{2.3cm}|}{$\textbf{G}=\{G^{t}, ..., G^{t+k}\}$} 
& \multicolumn{1}{|L{2.3cm}|}{$y \in \mathbb{R}^{|\mathbb{V}|}$}
\\
\multicolumn{1}{|L{4.6cm}|}{\textbf{\MTTReNN}}     
& \cmark 
& \cmark 
& \cmark 
& \multicolumn{1}{|L{2.3cm}|}{$\textbf{G}=\{G^{t}, ..., G^{t+k}\}$} 
& \multicolumn{1}{|L{2.3cm}|}{$y \in \mathbb{R}^{|\mathbb{V}| \times m}$} \\
            \bottomrule
        \end{tabular}
    }

    \vspace{0.35em}
    \begin{minipage}{0.95\linewidth}
        \footnotesize
        Legend: \cmark\ = supported;\; \xmark\ = not supported. \\[2pt]
    \end{minipage}
    \caption{
Comparison of neural network-based architectures for team modeling, organized by (i) their modeling capabilities (i.e., whether they capture relational information, temporal dynamics, or multiple tasks), and (ii) their input--output mapping.
\SNN cannot model relations, temporal dynamics, or multi-task predictions due to the structure of its input and output: the input $x \in \mathbb{R}^{d}$ represents a static team member (Def.~\ref{def:static_team_member}), which is mapped into a single team construct $y \in \mathbb{R}$ using a \SNN (Def.~\ref{def:snn}), as illustrated in Fig.~\ref{fig:snn}.
\TNN models temporal information, but cannot capture relations, since its input $x \in \mathbb{R}^{t \times d}$ only contains a temporal sequence for a single dynamic team member (Def.~\ref{def:static_team_member}). This input is mapped into a single team construct $y \in \mathbb{R}$ using a \TNN (Def.~\ref{def:tnn}; Fig.~\ref{fig:tnn}).
\ReNN captures relational structure because its input is a static team represented as a graph $G = (\mathbb{V}, \mathbb{E}, X)$ (Def.~\ref{def:static_team}). The model outputs a prediction for each team member, $y \in \mathbb{R}^{|\mathbb{V}|}$, using a \ReNN (Def.~\ref{def:tnn}; Fig.~\ref{fig:gnn}). However, it cannot capture temporal dynamics, since its information is limited to static features and relations.
\TReNN captures both relational and temporal aspects by modeling a dynamic team (Def.~\ref{dynamic_team}) as a sequence of temporal graphs $\mathbf{G} = \{G^{t}, \ldots, G^{t+k}\}$. It produces a single prediction for each team member over the entire temporal sequence, $y \in \mathbb{R}^{|\mathbb{V}|}$.
Finally, \MTTReNN extends \TReNN by enabling multi-task prediction. It uses the same dynamic-team input, but outputs multiple constructs per member, $y \in \mathbb{R}^{|\mathbb{V}| \times m}$, with $m$ denoting the number of tasks.
}
    \label{tab:comparison}
\end{table*}

\TNN models a \emph{dynamic team member} as a time-series of \emph{static team member} snapshots, as also explored in other works~\cite{sutskever2014sequence, avci2016predicting, 8626131,kim2016improving, lin2023interaction, 10191640} and illustrated in Fig.~\ref{fig:tnn}.

\begin{defn}
A \emph{dynamic team member} is defined as a pair $\{X, \textbf{y}\}$, where $X = (\textbf{x}^{1}, \textbf{x}^{2}, ..., \textbf{x}^{t})$ is a sequence of input features for a specific team member over $\textit{t}$ consecutive time steps and  $\textbf{y} \in \mathbb{R}^{m}$, where $m = 1$ is the output score referring to a team construct for the team member over the entire sequence.
\end{defn}

\begin{defn}
\label{def:tnn}
A \TNN uses its \emph{encoder} $f_{tnn}$ to map a sequence of input features $X \in R^{t \times d}$ into an embedding $h \in R^{m}$, and, similarly to Eq.~\ref{eq:decoder_traditional}, uses its \emph{decoder} $g(h)$  to produce a single output score $\textbf{y} \in \mathbb{R}^{m}$, where $m = 1$  for each team member. 
\begin{equation}
\label{eq:tnn}
(\textbf{h}^{t},\dots,\textbf{h}^{t+K}) = f_{\text{tnn}}(\{\textbf{x}^{t}, \textbf{x}^{t+1}, ..., \textbf{x}^{t+K}\})
\end{equation}
\end{defn}

However, \TNN models team members in isolation, missing the interaction information among team members, and ignores the possibility of jointly modeling multiple team constructs. 
Since the importance of interactions in teams has been widely studied~\cite{cannon1997teamwork, pype2018healthcare},
researchers have proposed team modeling approaches grounded in Graph Theory and Network Science~\cite{dyer1984team, salas1992toward, leenders2016once, chai2019applying, park2020understanding}, Complex Systems~\cite{dooley1997complex, arrow2000small, ramos2018teams}, or GNN~\cite{lin2020predicting, sharma2023graphitti}. We define these neural approaches as Relational Neural Networks (\ReNN), as represented in Fig.~\ref{fig:gnn}, and can be implemented as GCN~\cite{kipf2016semi}, Graph Attention Network~\cite{velickovic2017graph}, or Graph Isomorphism Network~\cite{xupowerful}.
They, indeed, can learn \emph{Social Embeddings} incorporating individual and interaction behavior via message passing strategies over the network representation of the (static) team, and open the road for relevant xAI-based applications~\cite{pmlr-v269-luca25a, azzolinreconsidering}, even though they have been largely ignored in the team modeling literature.
 
\begin{defn}
\label{def:static_team}
A \emph{static team} is a graph $G = (\mathcal{V}, \mathcal{E}, \textbf{X}, \textbf{Y})$, where $\mathcal{V}$ is the set of team members, $\mathcal{E}$ is the set of interactions, $\textbf{X} \in \mathbb{R}^{n \times d}$ are input features for each of the $n = |\mathcal{V}|$ team members, and $\textbf{Y} \in \mathbb{R}^{n \times m}$ are the corresponding outputs.
\end{defn}

\begin{defn}
\label{def:renn}
A \ReNN uses its \emph{encoder} $f_{renn}$ to encode each team member $v$ as $h_{v}$ and then, similarly to Eq.~\ref{eq:decoder_traditional}, uses its \emph{decoder} $g(h_{v})$ to map the embedding into the corresponding output score for the team member $y_{v} \in \mathbb{R}^{m}$, where $m = 1$:
\begin{equation}
\mathbf{h}_{v} = f_{\text{renn}}\left({\mathbf{x}_u \mid u \in \{{v} \cup \mathcal{N}(v)}\}\right)
\end{equation}
\end{defn}
where $\textbf{x}_{u}$ represents the features of team member $u$, while $\mathcal{N}(v)$ denotes the set of neighbors of $v$. 
Notice that this definition assumes a single processing step for simplicity. 
In standard \ReNN, this operation is recursively repeated to capture multi-hop relational dependencies, thereby enriching the \emph{social embedding} of each team member. \ReNN models interactions among team members.

\ReNN fails to encode dynamics, a crucial aspect in modeling team behaviour~\cite{marks2001temporally}. 
These limitations have motivated researchers 
to combine knowledge from dynamical systems and network science 
to model teams as adaptive complex systems~\cite{ramos2018teams}. This approach, however, struggles to process rich feature-based information, an area where deep learning architectures are particularly effective.

\subsection{Tempo-relational and multi-task modeling}
\label{sub:proposed_approaches}

To overcome the limitations of the previously mentioned architectures, we propose a novel approach for team modeling that leverages \TReNN that extends the view of teams as adaptive complex systems~\cite{ramos2018teams} by offering a data-driven, scalable approach that jointly accounts for temporal dynamics and interaction information, as represented in Fig.~\ref{fig:tgnn}.
 \TReNN can be implemented using EvolveGCN~\cite{pareja2020evolvegcn}, DySAT~\cite{sankar2020dysat}, or, in general, following the framework TGN~\cite{rossi2020temporal}.

\begin{defn}\label{dynamic_team}
A \emph{dynamic team} is defined as $T = (\textbf{G}, \mathbf{Y})$, where $\textbf{G} = \{G^{t}, ..., G^{t+K}\}$ is a sequence of $k$ time-indexed graphs $G^{j} = (\mathcal{V}^{j}, \mathcal{E}^{j}, \textbf{X}^{j})$, and $\textbf{Y} \in \mathbb{R}^{n \times m}$ are the corresponding outputs.
\end{defn}

\begin{defn}\label{def:tempo_relational_nn}
Given a \emph{dynamic team} as input, a \TReNN uses its \emph{encoder} $f_{trenn}$ to encode the temporal evolution of each team member $v$ into the embedding $h_{v}^{t:t+k}$, and uses its \emph{decoder}, similarly to Eq.~\ref{eq:decoder_traditional},  to map this embedding into the output space $y_{v} \in R^{m}$, where $m=1$. 

\begin{align}
\label{eq:trenn}
(\textbf{h}_{v}^{t},\dots,\textbf{h}_{v}^{t+K}) 
&= f_{\text{trenn}}\Big( \mathbf{x}_{u}^{t+i} | u^{t+i} \in \{v^{t+1} \cup \mathcal{N}(v^{t+i})\}, \nonumber \\
&\quad \forall i \in \{0,1, ..., K\}\Big) 
\end{align}

\end{defn}

Owing to the inherently interleaved nature of team constructs, we extend \TReNN with an MTL objective that enables the model to jointly represent and optimize multiple team constructs. This paradigm captures cross-task information and shared latent patterns (e.g., between emergent leadership and teamwork), leading to more robust and generalizable models of team functioning.

\begin{defn}
A \MTTReNN, similarly to Eq.~\ref{eq:trenn}, uses its encoder $f_{\text{trenn}}$ to map each team member $v$ of a \emph{dynamic team} into its corresponding embedding $h_{v}$, and uses its $m \geq 2$ \emph{multi-task decoders} $g_{\text{mt}}$ to map the embedding into multiple team constructs $\textbf{Y}_{v} \in \mathbb{R}^{m}$ for each team member in a team.
\end{defn}

\MTTReNN, as shown in Fig.~\ref{fig:mttgnn}, leverages shared parameters across tasks and task-specific decoders to jointly optimize all $m$ team constructs. This architecture enables more explainable and generalizable team modeling, particularly in high-stakes domains such as emergency response and healthcare. Rather than offering a mere incremental improvement, \MTTReNN represents a significant advancement from isolated predictions toward a holistic understanding of team dynamics.
Table~\ref{tab:comparison} compares the above mentioned architectures in terms of their modeling capabilities.

\begin{figure*}[htbp]
    \centering
    \begin{subfigure}{\textwidth}
        \centering
        \includegraphics[width=\textwidth]{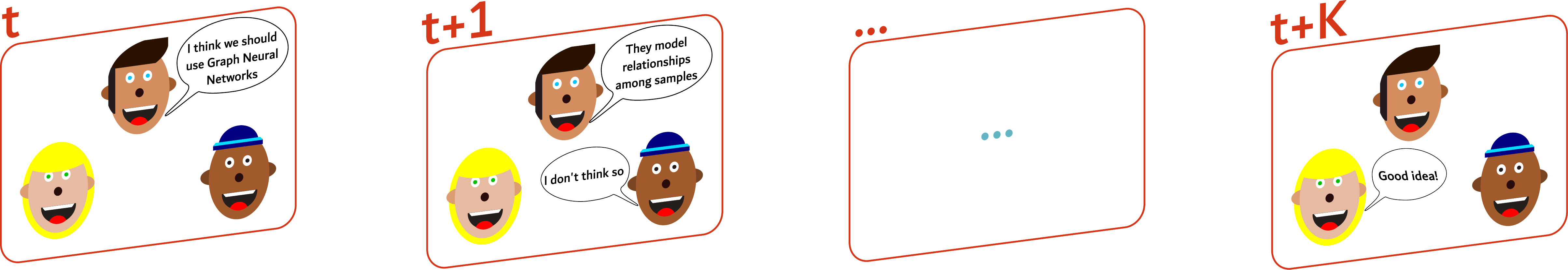}
        \caption{Team interaction}
        \label{fig:real_world_interaction}
    \end{subfigure}
    \begin{subfigure}{\textwidth}
        \centering
        \includegraphics[width=\textwidth]{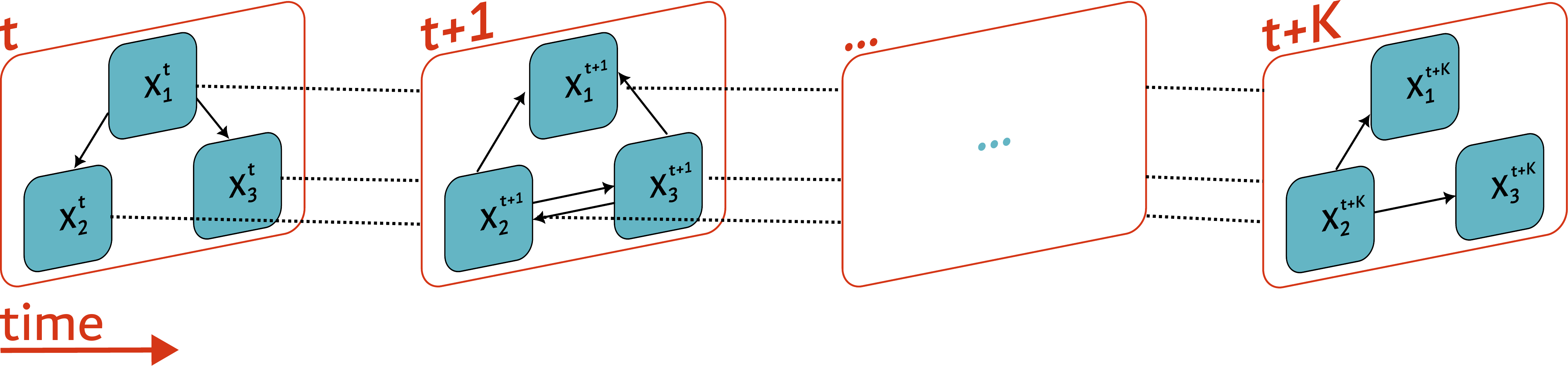}
        \caption{
        Snapshot temporal graph}
        \label{fig:temporal_graph}
    \end{subfigure}
    \begin{minipage}{\textwidth}
        \raggedright
        \includegraphics[width=\textwidth]{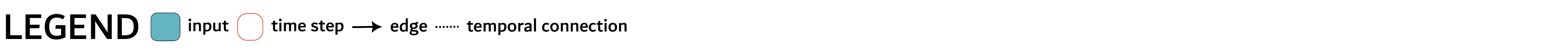}
        \label{fig:legend_team}
    \end{minipage}
    \caption{ 
   (a) A team interaction consists of a set of team members continuously interacting, exchanging verbal and nonverbal messages. Let us suppose that three team members are discussing the use of Graph Neural Networks to model team behavior. Initially (at time t), the team member with black hair tells the other team members that they should use Graph Neural Networks, later (at time t+1), the team member with the hat expresses his doubt about this idea. After some steps (at time t+K), finally, Lisa also states that she agrees with the others. This team interaction is converted by the temporal graph encoder into a snapshot temporal graph, as represented in (b). Each node $X_{i}^{t}$ represents the features of a team member (e.g., paralinguistic features), while each edge connecting a pair of nodes represents a one-to-one interaction (e.g., speaker-broadcast, where each team member who speaks has an edge connecting themselves to all the other team members).} 
    \label{fig:Dynamic_Team_Dataset_Mapping}
\end{figure*}

\section{Implementing \TReNN and \MTTReNN}
\label{sec:method}

In this section, we provide a detailed description of the \TReNN and \MTTReNN approaches.
We first present the input data (Sec.~\ref{sec:data_processing}), followed by the description of the encoder (Sec.~\ref{sec:encoder}), implemented as a \TReNN. Finally, we present the decoder (Sec.~\ref{sec:decoder}), which extends \TReNN into \MTTReNN, by leveraging MTL to jointly predict multiple team constructs (i.e., EL, LS, TW).

\subsection{From raw-data to temporal graph}
\label{sec:data_processing}

The raw input data, i.e., team behavioral data,  is processed by a temporal graph extractor. This extractor converts team interactions into temporal graphs through four stages: (i) temporal segmentation, (ii) subsegmentation, (iii) feature extraction, and (iv) topology generation.

The raw input data consists of a set of multimodal team raw data (e.g., audio, video) $\textbf{r}= (\textbf{r}_{1}, \textbf{r}_{2}, ..., \textbf{r}_{r})$, where each segment  $\textbf{r}_{i}$ corresponds to team interactions lasting $l_{i}$ seconds.
In the temporal segmentation stage, given an annotation frequency $f$, each team behavioral segment $\textbf{r}_{i}$ is partitioned into $ t_{i} = \lceil{\frac{l_{i}}{f}}\rceil$ subsegments, that we refer to as $\textbf{s}_{i}^{(k)}$, $\forall k \in \{1, 2, ..., t_{i}\}$. 
Each of these subsegments needs to be labeled with appropriate annotation $y_{i}^{k}$, $\forall k \in \{1, 2, ..., t_{i}\}$ that enables us to define a unit of information as the pair  $U_{i}^{(k)} = (\textbf{s}_{i}^{(k)}, y_{i}^{(k)})$.
In the subsegmentation stage, each segment $\textbf{s}_{i}^{(k)}$ is further separated into $z = \lceil\frac{m}{f}\rceil$ subsegments $\tilde{\textbf{s}}^{(k),j}_{i}$, each lasting $s$ seconds, such that: $\textbf{s}^{(k)}_{i} = \{\tilde{\textbf{s}}^{(k),1}_{i}, \tilde{\textbf{s}}^{(k),2}_{i}, ...,\tilde{\textbf{s}}^{(k),z}\}$.
Since annotations are performed at the team member level, the feature extraction stage computes the temporal features of each team member, while the topology generation stage derives their interactions over time. The combination of these two components, individual features and interactions, forms a \emph{dynamic team} as defined in Def.~\ref{dynamic_team} and represented in Fig.~\ref{fig:temporal_graph}. More specifically, given a subsegment $\tilde{\textbf{s}}^{(k),j}_{i}$, we obtain the corresponding static team $G_{i}^{(k), j} = \{V_{i}^{(k), j}, E_{i}^{(k), j}, X_{i}^{(k), j}\}$ . Then, a dataset composed of $p$ teams is denoted as $\textbf{T}=\{T_{1}, T_{2}, ..., T_{p}\}$. This pipeline shows how to automatically convert raw input data like the one in Fig.~\ref{fig:real_world_interaction} into a multivariate time series of features and interaction defined as a temporal graph.
This structure provides flexibility about the length of each team interaction and the topology of each graph, as the set of team members and their interactions can change over time.

\subsection{Temporal-graph encoding}
\label{sec:encoder}

The encoder maps input data into a robust and informative representation suitable for downstream predictions. 
Each sample from the dynamic team dataset (Sec.~\ref{sec:data_processing}) is a dynamic team, i.e., a sequence of temporally ordered team snapshots.
To model this structure, we instantiate our \TReNN\ architecture (Def.~\ref{def:tempo_relational_nn}) as a Space-Then-Time TGNN (STT-TGNN)~\cite{cini2023graph}.
The STT-TGNN decouples spatial (i.e., relational) modeling and temporal (i.e., dynamics) modeling. This decoupling is aligned with theories in team cognition~\cite{courtright2015structural}, which conceptualize team constructs as emerging from momentary relational configurations and their subsequent evolution over time. These two consecutive steps are realized as detailed below:
\begin{enumerate}
    \item Relational learning:  
    The relational learning module converts a \textit{dynamic team} into a sequence of tensors encoding relational information.     
    It receives as input a \emph{dynamic team} $T = (\textbf{G}, \mathbf{Y})$ (See Def.~\ref{dynamic_team}).
    Each dynamic team consists of a sequence of \emph{static team} snapshots $G^{j}$ captured at discrete time steps. 
    Each snapshot $G^{j}$ is independently encoded into an embedding tensor $H^{j}$. 
    \item Temporal learning: The temporal learning module converts the output of the relational learning module, i.e., a sequence of tensors encoding relational information $\textbf{H}_{i}$, into a tensor embedding capturing also the temporal evolution of the team.
\end{enumerate}

In the following subsections, we formalize the structure of the \emph{encoder} according to the STT-TGNN implementation: (i) the relational learning module; (ii) the temporal learning module; and (iii) the contrastive learning module, which further enhances the robustness of the embeddings.

\subsubsection{Relational learning}
The relational learning module is based on Message Passing Neural Networks (MPNNs)~\cite{gilmer2017neural}, a well-known paradigm to model non-sequential and variable-size relations among input data, as illustrated in Fig.~\ref{fig:final_gnn}.
MPNNs propagate information across nodes through an iterative updating mechanism. It encodes information from neighboring nodes and combines it with the hidden representation of the current node.
{Given a dynamic team, i.e., a sequence of \emph{static team} snapshots $G^{j}$,
the relational learning module processes each static team snapshot independently $G_{i}^{(k), j} = \{\mathcal{V}_{i}^{(k), j}, \mathcal{E}_{i}^{(k), j}, X_{i}^{(k), j} \}$.
For each team member $v \in \mathcal{V}_{i}^{(k),j}$, the relational learning module operates through two learnable components:
\begin{itemize}
    \item a message function computing a message $m$ from the neighboring nodes of the current node $\forall u \in \mathcal{N}(\textbf{v})$ and   
    their corresponding edge-features $\textbf{e}_{vu} \in \mathcal{E}_{i}^{(j),k}$:
    \begin{equation}
    \textbf{m}_{v} = \sum_{u \in \mathcal{N}(v) } m\left(\textbf{h}_{v}, \textbf{h}_{u}, \textbf{e}_{vu}\right)
    \end{equation}
    
    \item an update function generating an updated node hidden representation $h^{'}_{v}$ as the combination of the message from the neighboring nodes $m_{v}$ with the current node hidden representation $h_{v}$:

    \begin{equation}
    \textbf{h}^{'}
    _{v} = u\left(\textbf{m}_{v}, \textbf{h}_{v}\right)
    \end{equation}
\end{itemize}
Under the MPNNs framework, both the message function and the update function are differentiable and learnable functions. We model the kernel of this function as a GCN~\cite{kipf2016semi}, which generalizes the convolution operator in order to graph data to compute the hidden representation for all the team members according to the following formula:
\begin{equation}
H^{t} = \phi\left(\tilde{D}^{-\frac{1}{2}}\tilde{\textbf{A}}\tilde{D}^{-\frac{1}{2}}X^{\textit{t}}\textbf{W}\right)
\end{equation}
where  $\phi$ is an activation function, $t$ is the current time step, $H^{t}$ refers to the hidden representation of all the team members at the current time step, $\tilde{\textbf{A}}$ is the adjusted adjacency matrix with added self-connection to increase stability, $\tilde{D}$ is the degree of the adjusted adjacency matrix, $X^{t}$ is the tensor containing individual speech features of all team members at the time step $t$ and $\textbf{W}$ refers to the learnable convolution parameters.\\

\begin{figure*}
    \centering
    \begin{tabular}{c c c}
        \includegraphics[width=0.5\textwidth]{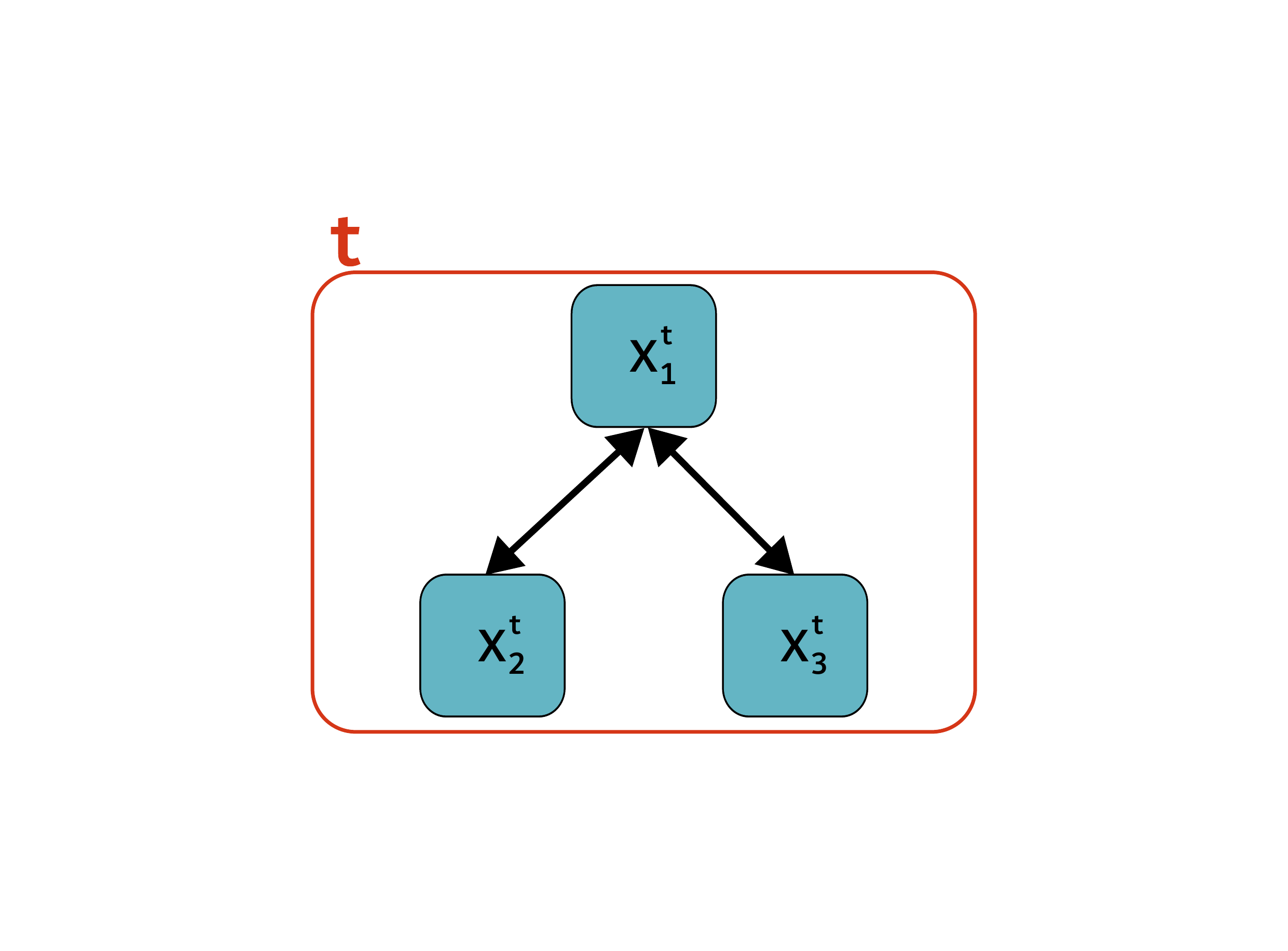} & 
        \includegraphics[width=0.5\textwidth]{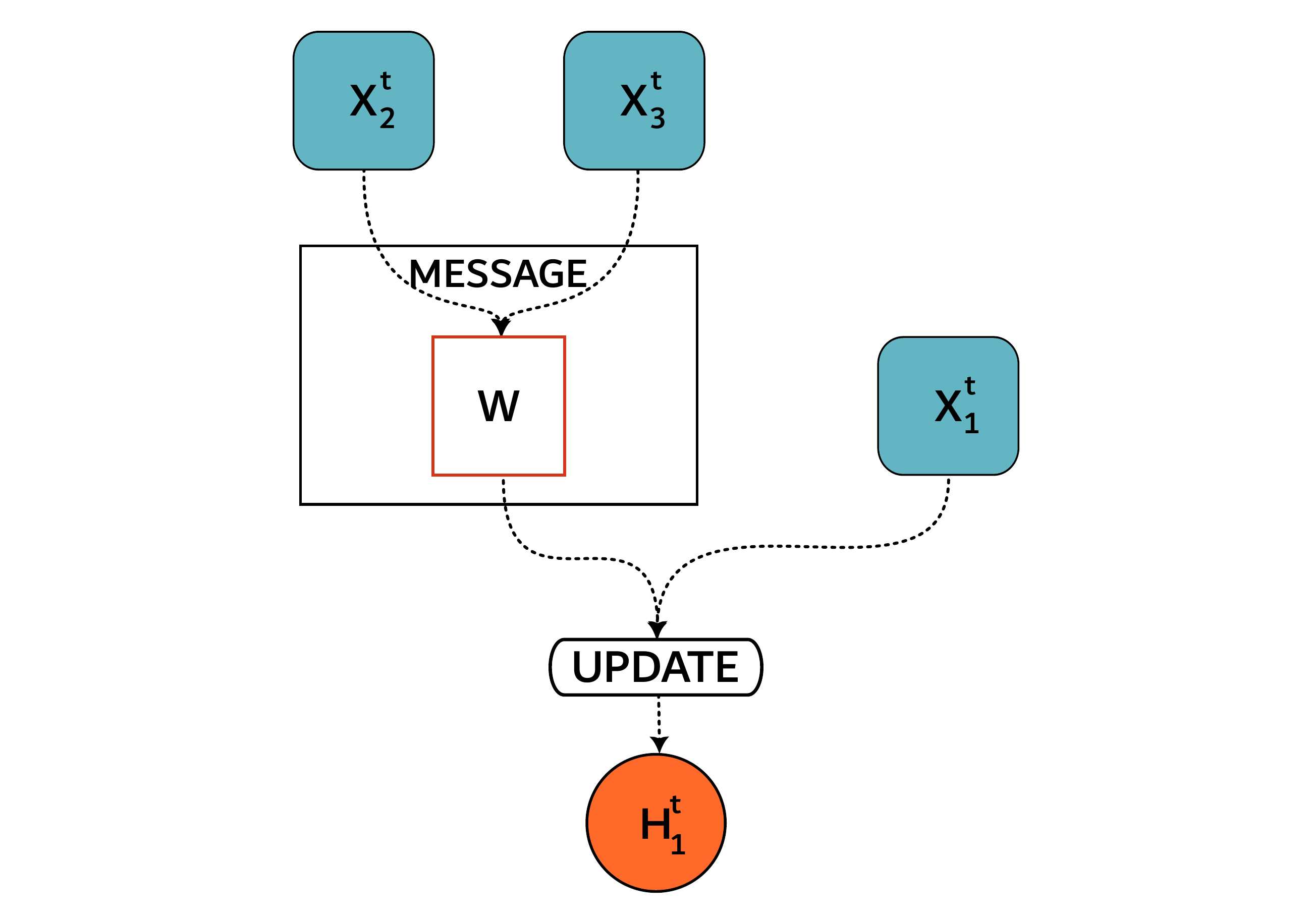}  \\
        (a) Temporal graph snapshot & (b) Message Passing procedure for a node \\
    \end{tabular}
    \begin{minipage}{\textwidth}
        \raggedright
        \includegraphics[width=\textwidth]{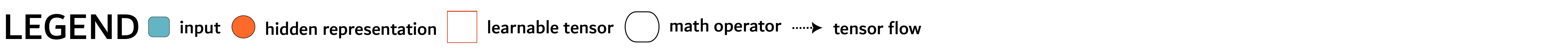}
        \label{fig:legend_mp}
    \end{minipage}
    \caption{The MPNNs' steps to map topological structure into a single tensor. (a) Let us consider  
    a snapshot graph at time $t$ with three nodes ($X_{1}^{t}, X_{2}^{t}, X_{3}^{t}$) and two bidirectional edges. This topology encloses both individual node information (i.e., features) and topological information (i.e., edges). (b) Generation of the hidden representation (i.e., $H_{1}^{t}$) for the first node (i.e., $X_{1}^{t}$). Since the node is connected by an edge to two neighboring nodes (i.e., $X_{2}^{t}, X_{3}^{t}$), the individual features of these two nodes are combined into a single message through the \emph{Message} module. The resulting message is combined with the original representation of the first node (i.e., $X_{1}^{i}$) through the \emph{Update} module, resulting in the final hidden representation of the first node $H_{1}^{t}$.} 
    \label{fig:final_gnn}
\end{figure*}

\subsubsection{Temporal learning}
The temporal learning module is based on MHA~\cite{vaswani2017attention} to effectively capture long-range dependencies across time steps.
MHA receives as input the output of the relational learning module. At this stage, each dynamic team is encoded into a sequence of team member embeddings $\textbf{H}^{t} \in R^{n \times d}$ for each time step $t$. 
Each row of these embeddings  $\textbf{H}_{i}^{t} \in R^{d}$ encodes the behavioral features and the interactions of a team member but it still completely ignores time dependencies. 
\begin{figure}
    \centering
    \includegraphics[width=0.4\textwidth]{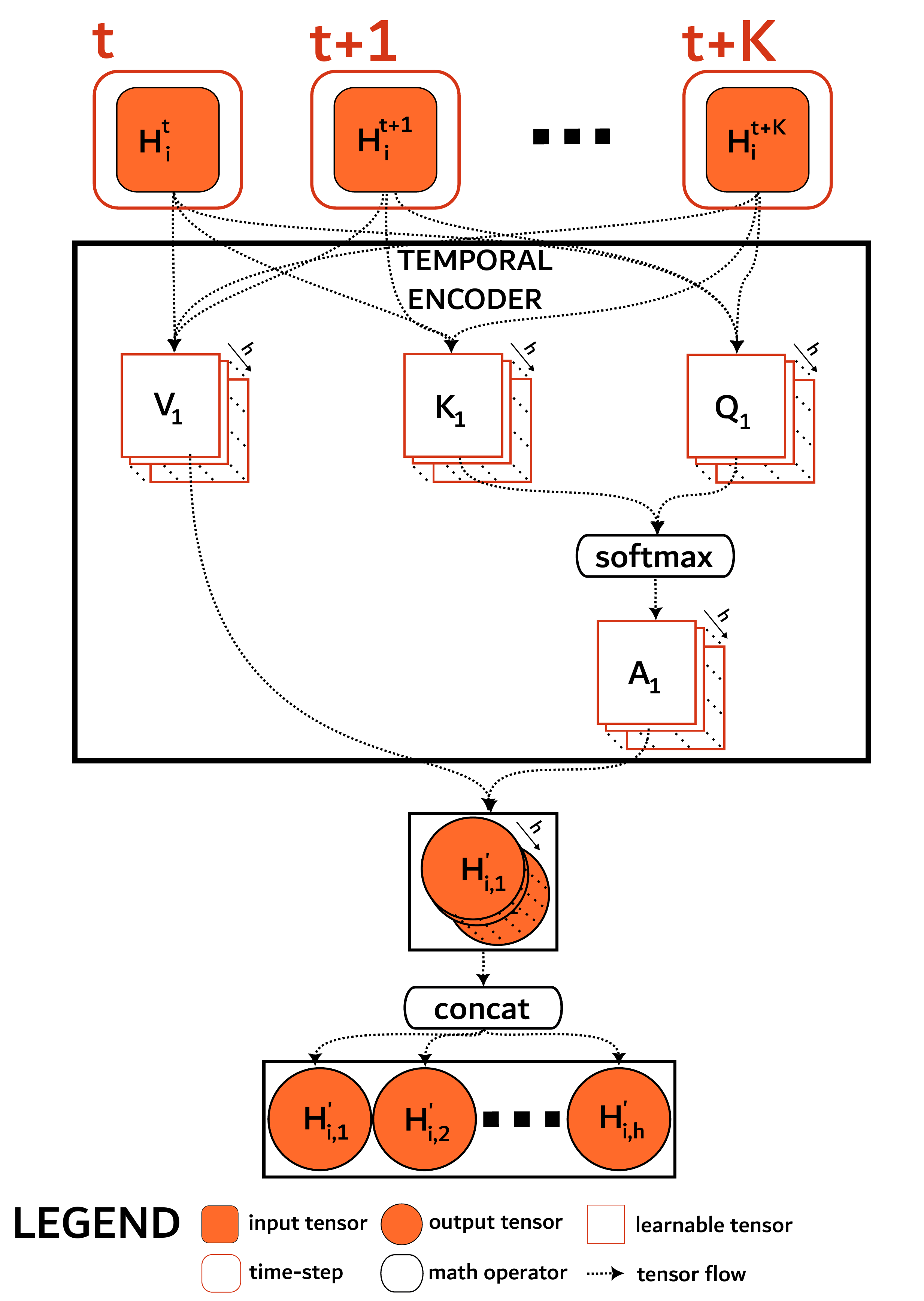}
    \caption{
    Given the hidden representation of a team member $i$ at time $t$, $H^{t}_{i}$, until time-step $t+K$, the input sequence \{$H_{i}^{t}$, $H_{i}^{t+1}, ..., H_{i}^{t+K}$\} is represented by the orange boxes that are the inputs to generate the query tensor $Q$ and the key tensor $K$ (the white boxes). Next, $Q$ and $K$ are combined to generate the Attention tensor $A_{j}$ containing the attention scores. Then, the sequence of attention tensors $A$ is combined with the value tensor $V$ to generate the resulting hidden representation encoding temporal information} $H^{'}=\{H^{'}_{i,1}, H^{'}_{i,2}, ..H^{'}_{i,h} \}$. 
    \label{fig:multi_head_attention}
\end{figure}
To model time dynamics, MHA is introduced, it is built on top of the self-attention mechanism, which measures how much a specific time-step in a sequence is relevant to model another time-step in the same sequence, as illustrated in Fig.~\ref{fig:multi_head_attention}.
For each team member, three learnable projection matrices are defined: the query-weight tensor $W_{q} \in \mathbb{R}^{d \times d_{K}}$, the key-weight tensor $W_{k} \in \mathbb{R}^{d \times d_{K}}$, and the value-weight tensor $W_{v} \in \mathbb{R}^{d \times d_{v}}$. Each one of them, respectively, is combined with the input $H_{i}$ to generate the projected representations computed as:
\begin{equation}
\small
Q_i = H_i W_q \in \mathbb{R}^{K \times d_k},\;
K_i = H_i W_k \in \mathbb{R}^{K \times d_k},\;
V_i = H_i W_v \in \mathbb{R}^{K \times d_v}
\end{equation}

where $K_{i} $ is the key tensor, $ Q_{i}$ is the query tensor, and $V_{i}$ is the value tensor, while $K$ is the sequence length, $d_{K}$ is the key and query tensor dimension, and $d_{v}$ is the value tensor dimension. Once these matrix multiplications are computed, each self-attention map $A_{i} $ is computed as reported in Eq.~\ref{eq:attention_weights}. 
\begin{equation}
\label{eq:attention_weights}
A_{i}= \textit{softmax}\left(\frac{Q_{i} K_{i}^{\top}}{\sqrt{d_{k}}}\right)  \in R^{K\times K} \; ,  \forall i \in \{1, 2, ..., h\}
\end{equation}
In Eq.~\ref{eq:attention_weights} the attention scores are converted into attention probabilities via \textit{softmax}, and they can be combined with the value tensor \textbf{$V_{i}$} to get the hidden representation as reported in Eq.~\ref{eq:head_computation}.
\begin{equation}
\label{eq:head_computation}
\textbf{H}^{'}_{i} = A_{i} \times V_{i} \in R^{K \times d_{v}}\; ,  \forall i \in \{1, 2, ..., h\}
\end{equation}
As previously mentioned, MHA extends the representational capacity of self-attention by applying multiple attention mechanisms in parallel, each with its own set of learnable parameters, and combines them via concatenation as reported in Eq.~\ref{eq:MHA}:
\begin{equation}\label{eq:MHA}
\text{MHA}(Q, K, V) = \textit{Concat}(\textbf{H}^{'}_{1},\textbf{H}^{'}_{2}, ..., \textbf{H}^{'}_{h}) 
\end{equation}
Thus, the model focuses on different temporal aspects of the sequence, improving its ability to capture complex and long-range dependencies in variable-length sequences.

\subsubsection{Contrastive learning}

Beyond architectural design, a common challenge in team modeling is the scarcity and noisiness of labeled data, which stems from both the subjectivity of human annotations and the complexity of the high-level constructs being labeled. 
While data augmentation techniques are commonly used to mitigate the limited availability of data, they alone are not sufficient to ensure effective generalization, especially when the labels themselves are noisy and partially subjective.

To address this limitation, we introduce contrastive learning, which operates directly at the loss level to shape the structure of the embedding space. Rather than relying only on input-level transformations, we encode task-relevant inductive biases into the learning dynamics by incorporating ordinal supervision derived from human assessments.
Each team member receives a real-valued score reflecting their perceived role or contribution. From these scores, we derive a within-team ranking: a relative ordering that expresses the team’s internal structure. While standard supervised losses typically ignore such relational information, we leverage it via a Pairwise Ranking Loss that promotes embeddings to be aligned with the annotated hierarchy, as shown in Fig.~\ref{fig:pairwise_loss}.
Given a ranking function $f(x)$ and two team members $x_i$ and $x_j$ with $x_i$ ranked higher than $x_j$, the loss is defined as:
\begin{equation}
L_{\text{p}} = \sum_{(i, j) \in D} \max(0, 1 - (f(\mathbf{x}_i) - f(\mathbf{x}_j))),
\end{equation}
where $D$ is the set of correctly ordered pairs of team members. This hinge-based loss enforces a margin between embeddings, encouraging the model to preserve intra-team ranking structure in the latent space.
This contrastive signal complements traditional supervised objectives and induces a structured embedding geometry that reflects the underlying team hierarchy, thereby enhancing the consistency, robustness, and cross-task generalization of downstream predictions.

\begin{figure}
    \centering
    \includegraphics[width=0.5\textwidth]{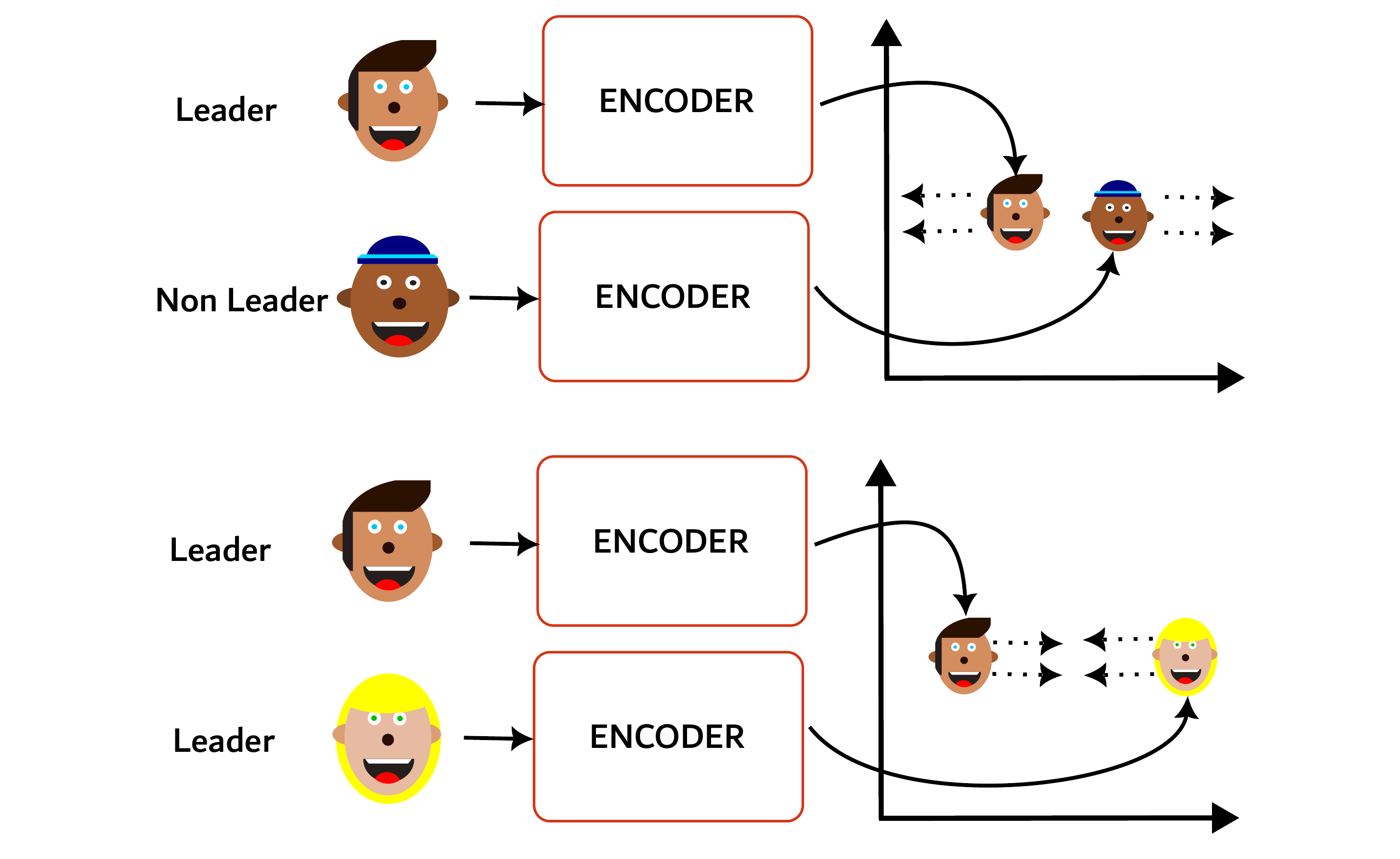}
    \caption{    Pairwise Ranking Loss enforces embeddings of the leaders to be closer in the embedding space, while the embedding of non-leaders to be farther from the ones of the leaders.}
    \label{fig:pairwise_loss}
\end{figure}

\subsection{From encoding to predictions}
\label{sec:decoder}
\begin{figure*}[htbp]
    \centering
    \begin{subfigure}[t]{0.48\textwidth}
        \includegraphics[width=\textwidth]{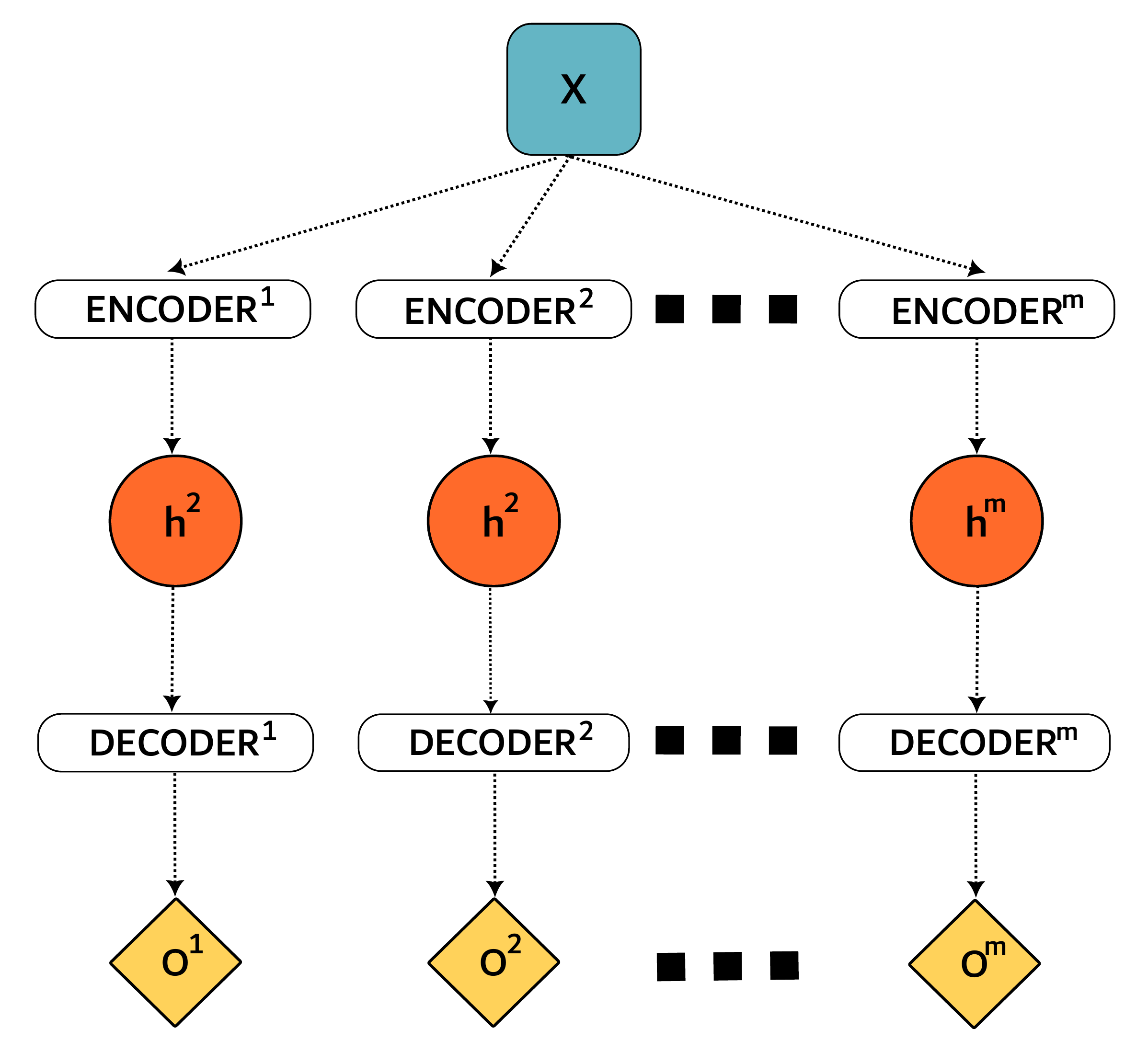}
        \caption{Traditional learning}
        \label{fig:traditional_mlp}
    \end{subfigure}
    \hfill
    \begin{subfigure}[t]{0.48\textwidth}
        \includegraphics[width=\textwidth]{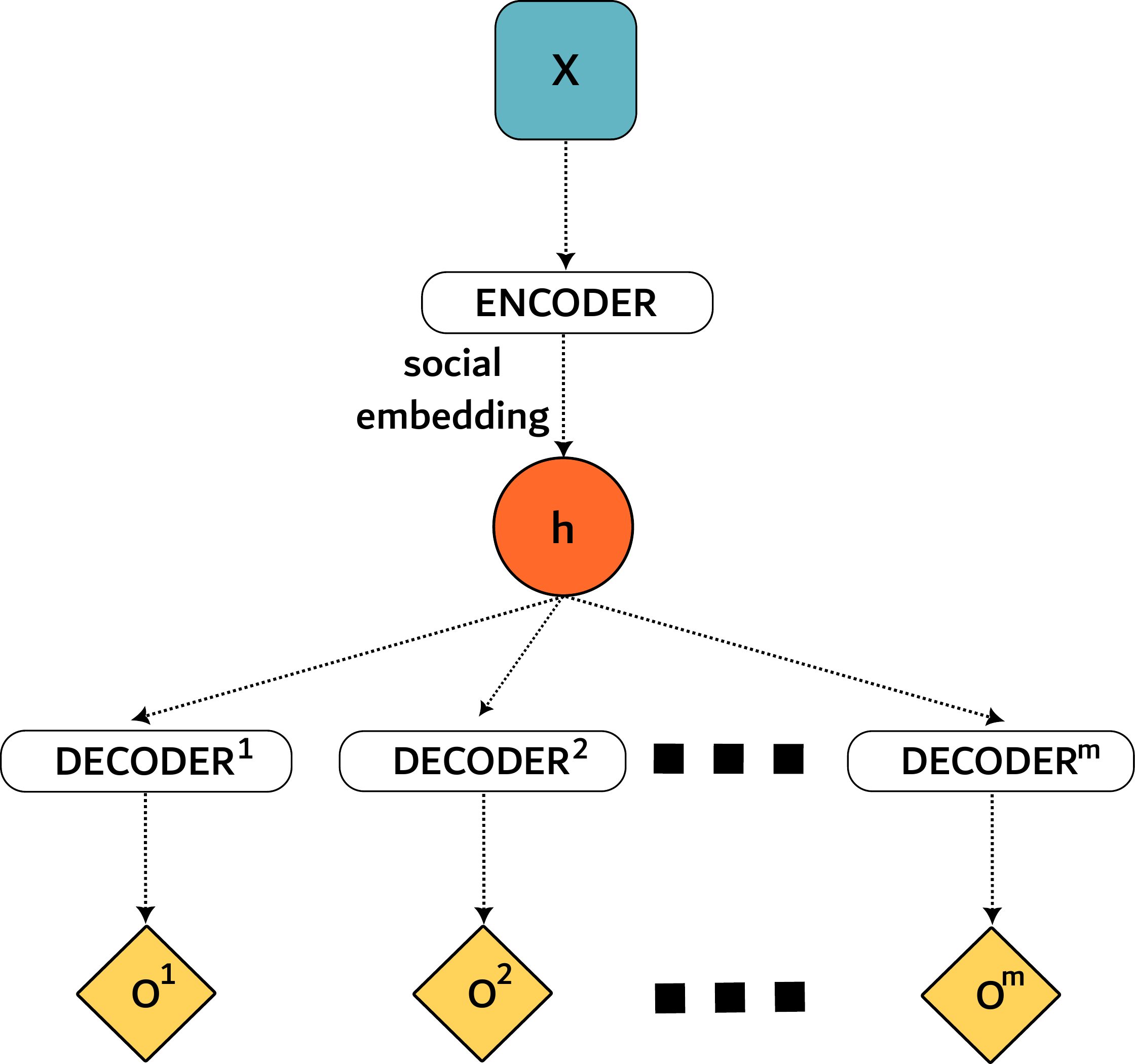}
        \caption{Multi-task learning}
        \label{fig:multi_task_learning}
    \end{subfigure}
    \vspace{1em}
    \begin{minipage}{\textwidth}
        \raggedright
        \includegraphics[width=\textwidth]{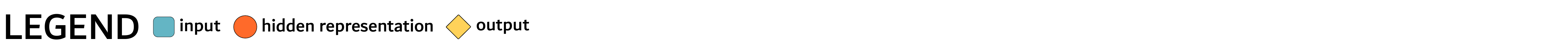}
        \label{fig:legend_mtl}
    \end{minipage}
    \caption{
    Two principal approaches for modeling multiple tasks in Neural Networks. (a): Traditional learning approach: the input (blue node) 
    feeds multiple backbones having different parameters. The output of the backbones are embeddings (orange nodes) that feed linear layers mapping them into the output space (yellow node) given an input sample (blue square) composed of three features, two parallel encoders composed of two hidden layers having hidden size four and three (orange square), learns different parameters to feed it as input to two different linear layers mapping these hidden representation into the output space (yellow rhombi). (b) Multi-task Learning approach: the input (blue node) feeds a shared backbone that outputs a Social Embedding, an embedding encoding features to solve multiple tasks. The output, indeed, feeds two different specialized heads that learns specific parameters to specialize the embedding into different output spaces (yellow node).}
    \label{fig:MultiTaskLearning}
\end{figure*}

When the task is predicting a single team construct, \TReNN is employed, and the decoder reduces to a single FFNN. In the following, we focus on the more complex case when multiple team constructs are jointly modeled via the \MTTReNN architecture. In this architecture, the decoder is based on the MTL paradigm~\cite{zhang2021survey}, which jointly learns multiple tasks (i.e., team constructs) rather than treating them as independent tasks. MTL employs a shared encoder across all the team constructs to map the input \emph{dynamic graph} to a shared \emph{social embedding} space (Def.~\ref{def:social_embedding}) for each team member. This embedding is then passed to task-specific heads, each predicting a different team construct.
This design allows the model to simultaneously capture shared dynamics across tasks while maintaining specialization for individual constructs (Fig.~\ref{fig:MultiTaskLearning}).
Formally, given a set of tasks $\mathcal{A} = \{A_{1}, A_{2}\dots, A_{m}\}$, with $m \geq 2$,  the resulting MTL loss is defined as a weighted sum of task-specific losses:
\begin{equation}
\mathcal{L}_{\text{MTL}} = \sum_{i=1}^{m} \lambda_i \mathcal{L}_{T_i}
\end{equation}
where each weight $\lambda_i$ reflects the relative importance of the corresponding team construct $T_i$. To ensure positive coefficients $\lambda_i > 0$ for all $i$, we parameterize them via an exponential transformation:
\begin{equation}
\lambda_i = \exp(\alpha_i)
\end{equation}
with $\alpha_i \in \mathbb{R}$ being optimized during the training procedure. This approach ensures no task is ignored during the joint optimization.

Unlike traditional learning (Fig.\ref{fig:traditional_mlp}), in which different team constructs do not share any parameters, and unlike fine-tuning, in which each decoder is optimized \emph{post-hoc} independently for each task requiring an additional training procedure, in MTL (Fig. \ref{fig:multi_task_learning}), the backbone is trained jointly with all the heads, allowing mutual refinement of the shared encoder and task heads. 

MTL exhibits two major advantages. First, it acts as a regularizer: joint optimization across tasks acts as an upper bound on the weights and the variance, resulting in the prevention of overfitting. Second, it improves efficiency, reducing both time complexity due to the single training procedure for all the tasks and memory cost due to the shared encoder parameters. 

\section{Experiments and Results}
\label{expres}
In this section, we first present our research questions. 
Next, we detail the experimental setup, covering datasets, evaluation protocol, and implementation details. Finally, results are reported and discussed. 
Our work aims to address the following research questions:
\begin{itemize}
 \item[\textbf{Q1:}] Does \TReNN outperform existing paradigms in team modeling?
\item[\textbf{Q2:}] Does joint learning of Multiple Tasks increase efficiency in team modeling? 
\item[\textbf{Q3:}] Does \MTTReNN provide insights about team member contribution to team dynamics?
\end{itemize}

\subsection{Datasets and annotations}
\label{sec:datasets}
We used two state-of-the-art ream datasets: PAVIS~\cite{beyan2016detecting} and  ELEA~\cite{sanchez2011audio, sanchez2012nonverbal}. Both datasets were recorded in laboratory settings and are based on the Winter or the Summer Survival task, a well-established team-decision-making task requiring teams to rank objects by importance for survival in extreme weather conditions.
PAVIS is an audio-video dataset consisting of 16 teams of four people.
Each team had a designated leader, appointed based on preliminary aptitude questionnaires filled out by participants.
ELEA is an audio-video dataset of 40 teams of three or four peers. No specific roles, indeed, were attributed to the team members. 
In this study, we used a standard subset of these datasets consisting of 12 teams from PAVIS and 27 teams from ELEA, respectively. The other teams were discarded due to technical reasons (e.g., missing audio). The duration of recordings ranges between 10 and 30 minutes: ELEA averages 14.6 minutes (SD=1.68), and PAVIS 21.6 minutes (SD=2.24), further details about the datasets are reported in Table~\ref{tab:datasets}.

\begin{table}[t]
\centering
\renewcommand{\arraystretch}{1.15}
\setlength{\tabcolsep}{10pt}
\begin{tabular}{p{3.2cm} p{1.75cm} p{1.75cm}}
\toprule
\textbf{Characteristic} & \textbf{PAVIS} & \textbf{ELEA} \\
\midrule
Scenario 
& Survival task
& Survival task \\
\# Teams 
& 12
& 27\\
\# Members per team 
& 4
& 3-4 \\
Total duration 
& $\sim$5 hours 
& $\sim$7 hours \\
Video recording
& 4 cameras 
& 2 cameras \\
Audio recording
& 4 microphones 
& 1 microphone\\
Annotation granularity 
& 6 minutes
& 6 minutes \\
Emergent Leadership label
&  GLIS
& GLIS$^{*}$ \\
Leadership style label & SYMLOG$^{*}$ &  SYMLOG$^{*}$ \\
Teamwork label &  BFTQ$^{*}$ & BFTQ$^{*}$ \\
\bottomrule
\end{tabular}
\caption{Summary of the characteristics of the PAVIS and ELEA datasets. $*$ indicates that the labeling procedure has been implemented by four annotators carried out directly for this work.}
\label{tab:datasets}
\end{table}

Annotations were obtained from four independent raters, each scoring different video subsets. The annotation granularity was set at 6-minute intervals.
To operationalize the team constructs of interest, raters evaluated team functioning using validated instruments that measure core aspects of TW, EL, and LS. Specifically, the annotations encompass:
\begin{itemize}
\item TW components according to the BFT questionnaire~\cite{van2012development}: It covers the five core behavioral components (see Table~\ref{tab:main_tw_components}) and the three coordinating mechanisms (see Table~\ref{tab:coordinating_mechanisms}) of the BFT framework. The questionnaire has 37 items assessed on a 7-point Likert scale: 5 items for A, MPM, and CC; 3 items for BB; 4 items for TL and SMS; 9 items for TO, and 2 items for MT; 
\item the General Leader Impression Scale (GLIS)~\cite{lord1984test} questionnaire, consisting of 5 items rated on a 5-point Likert scale;
\item the Systematic method for the Multiple Level Observation of Groups (SYMLOG)~\cite{bales1980symlog,koenigs2002symlog} questionnaire, a 26-item questionnaire assessed on a 5-point Likert scale referring to LS through three bipolar dimensions: Dominance vs. Submissiveness, Friendliness vs. Hostility, and Task-orientation vs. Expressiveness.
\end{itemize}

\subsection{Data pre-processing}
The experiments we conducted, using \TReNN and \MTTReNN implemented as an STT-TGNN (see Sec.~\ref{sec:encoder}), take as input the datasets described in Sec.~\ref{sec:datasets} from which paralinguistics features were extracted.
The graph topology is defined through speaker diarization~\cite{radford2023robust}, the first step of pre-processing, that determines "who speaks when". A team member speaking is defined as a message-sender, while all the other team members as receivers. Such information is encoded in the graph through a directed edge from the node corresponding to the sender to all the other team members.
Then, for each speaking team member, the 88 features of the eGeMAPS speech feature set~\cite{eyben2015geneva} were extracted through the open\-SMILE library~\cite{eyben-2010}. These features are paralinguistic features related to prosody (e.g., pitch, loudness), voice quality (e.g., jitter, shimmer), and spectral characteristics, validated by speech scientists, psychologists, and computer scientists.

Statistical functionals such as mean, normalized standard deviation, and percentiles are computed over the feature set to summarize their temporal dynamics within each time window.\\

\subsection{Evaluation protocol}
We adopted a nested \textit{Leave-One-Group-Out} (LOGO) evaluation protocol. Given the set of all available teams \(\mathbf{T} = [T_{1}, T_{2}, \ldots, T_{n}]\), at each iteration one team \(T_i\) is selected as the validation set, and a different team \(T_j\) (\(j \neq i\)) as the test set. The remaining teams denoted as $T^{\mathrm{trn}}$ form the training set, i.e.,:
\[
T^{\mathrm{trn}} = \mathbf{T} \setminus \{T_{i} \cup T_{j}\}.
\]
This procedure is repeated for all possible combinations of validation-test team pairs, and the final performance is reported as the average over all resulting evaluations.
Formally, let \(P(f(\cdot), T_j)\) denote a performance metric of the model \(f\) on the test team \(T_j\), and \(f(\mathbf{T} \setminus \{T_i \cup T_j\}, T_i)\) be the model trained on \(\mathbf{T} \setminus \{T_i \cup T_j\}\) and validated on \(T_i\). Then, the expected performance is defined as:
\begin{equation}
    \mathbb{E}[P(\mathbf{T})] = \frac{1}{n(n-1)} \sum_{i=1}^{n} \sum_{\substack{j=1 \\ j \neq i}}^{n} P\big(f(\mathbf{T} \setminus \{T_i \cup T_j\}), T_j \big)
\label{eq:leave_one_out}
\end{equation}

The following evaluation metrics were adopted:
\begin{itemize}
    \item Mean Squared Error (MSE): computed as the average squared difference between predicted and ground truth team construct scores across all team members $e$ in the test set $T_{j}$:
    \begin{equation}
    \mathbb{E}[\mathrm{MSE}(f, T)] = \frac{1}{n(n-1)} \sum_{i=1}^{n} \sum_{\substack{j=1 \\ j \neq i}}^{n} \sum_{e \in T_j} \big(f(e) - e_M\big)^2
    \end{equation}
    where \(f(e)\) is the predicted score and \(e_M\) the ground truth for team member member \(e\).
    \item Accuracy@1 and Accuracy@Last: measuring whether the model correctly identifies the highest- and lowest-ranked team members in the test team $T_{j}$ according to a specific team construct score of interest:
    \begin{equation}
    \label{eq:acc_one}
    ACC@1 = \mathbbm{1}\left( \arg\max_i M(T_j) = \arg\max_i f(T_j) \right)
    \end{equation}
    \begin{equation}
    \label{eq:acc_last}
    ACC@Last = \mathbbm{1}\left( \arg\min_i M(T_j) = \arg\min_i f(T_j) \right)
    \end{equation}
\end{itemize}

These metrics highlight both the prediction accuracy as well as the model’s ability to preserve ordinal rankings within each team, and are measured over all the teams according to the LOGO protocol described above.

\begin{table*}
    \centering{ 
        \begin{tabular}{l|l|cc|cc|}
        \toprule
        & & \multicolumn{2}{c|}{\textbf{ACC@1 - Emergent Leadership}} & \multicolumn{2}{c|}{\textbf{ACC@Last - Emergent Leadership}} \\
        \textbf{Paradigm} & \textbf{Model}   & PAVIS & ELEA & PAVIS & ELEA \\
        \midrule
        \multirow{2}{*}{\centering \SNN} 
            & \col{RF}  &  \col{\mstd{79.54}{0.18}} & \col{\mstd{70.28}{0.12}} & \col{\mstd{78.97}{0.53}} & \col{\mstd{71.28}{0.47}} \\
            & FFNN  & \mstd{77.05}{0.54} &  \mstd{67.68}{0.94} & \mstd{78.24}{0.72} & \mstd{70.11}{0.65} \\
        \midrule
        \multirow{2}{*}{\centering \TNN} 
            & \col{LSTM}  &  \col{\mstd{80.69}{0.72}} & \col{\mstd{71.39}{0.72}} & \col{\mstd{82.06}{0.68}} & \col{\mstd{72.29}{0.58}} \\
            & MHA   &  \mstd{81.25}{0.56} & \mstd{72.48}{0.52} & \mstd{83.37}{0.45} & \mstd{74.98}{0.44} \\
        \midrule
        \multirow{1}{*}{\centering \ReNN} 
            & \col{GCN}   &  \col{\mstd{83.96}{0.64}}  & \col{\mstd{73.18}{0.76}} & \col{\mstd{83.96}{0.64}} & \col{\mstd{74.22}{0.48}} \\   
        \midrule
        \multirow{1}{*}{\centering \TReNN} 
            & \col{GCN+MHA}  & \col{\mstd{\textbf{85.95}}{0.48}} & \col{\mstd{\textbf{74.55}}{0.48}} & \col{\mstd{\textbf{85.95}}{0.58}} & \col{\mstd{\textbf{75.42}}{0.40}} \\
        \midrule      
        \bottomrule
    \end{tabular}
    }
    \caption{The average ACC@1 and ACC@Last of EL over 10 different seeds. The significantly best-performing model is boldfaced. Different models are grouped according to their main modeling paradigm (i.e., \SNN, \TNN, \ReNN, \TReNN)}
    \label{tab:emergent_leadership}
\end{table*}

\subsection{Experimental results}

In the following subsections, we investigate our three research questions.

\paragraph{{\bf Q1: Does \TReNN outperform existing paradigms in Team modeling?}}

To assess the effectiveness of \TReNN for team modeling, we evaluated it on the task of predicting EL, LS, and the TW components (see Table~\ref{tab:main_tw_components} and Table~\ref{tab:coordinating_mechanisms}).\\ 

\paragraph{\textit{Emergent Leadership}}
EL prediction is formulated as a regression task in which paralinguistic features are the predictors, and the predicted variable is a score of EL for each team member in a team. EL scores are computed as the average of the scores obtained by GLIS.  
Table~\ref{tab:emergent_leadership} reports the mean and standard deviation, over 10 different seeds, of ACC@1 and ACC@Last scores across the models described in Sec.~\ref{sec:snn_to_trenn}: Random Forest (RF) and FFNN as SNNs, LSTM and MHA as TNNs, GCN as \ReNN. As instances of \TReNN, the combination of GCN to model relations and MHA to model time direction was used as described in Sec.~\ref{sec:method}. RF is implemented using 70 trees. The FFNN is designed as a three-layer architecture with the embedding size set to 16 with a weight decay of 0.01 for both ELEA and PAVIS to mitigate overfitting, while the learning rate is set to 0.0025 for PAVIS and 0.00125 for ELEA. The LSTM model is configured as a four-layer architecture with the embedding size set to 32, a weight decay of 0.015, and a learning rate of 0.001 for PAVIS and 0.0075 for ELEA. The MHA model employs four attention heads within a four-layer architecture with embedding size set to 64, with a weight decay of 0.012 for PAVIS and 0.02 for ELEA, and a learning rate of 0.00125 for both datasets. The GCN is implemented as a three-layer architecture with embedding size set to 32, a weight decay of 0.005, and a learning rate of 0.0075 in the case of PAVIS, while weight decay is set to 0.003 and learning rate 0.01 for ELEA. Finally, \TReNN combines a two-layer GCN and a two-layer MHA with four attention heads with embedding size to 64, using a learning rate of 0.006 and a weight decay of 0.009 in the case of PAVIS and learning rate is set to 0.005.
We assessed potential significant differences between the performances of the models through statistical tests with a significance $\alpha$ level equal to 0.05. First, tests to compare pairs of models implementing the same paradigm were run. Then, a between-paradigms test was performed with a post hoc analysis when needed and adjusting the p-value using the Bonferroni correction. 
The Shapiro-Wilk test showed a significant departure from normality of data. Thus, we chose non-parametric paired tests. 

We conduct a two-sided Wilcoxon test to compare model pairs within each paradigm.
For the SNN paradigm, the test shows that there is a significant difference between RF and FFNN on all metrics and datasets (all $p < 0.05$). Similarly, about the TNN paradigm, MHA performs significantly better than LSTM (all $p < 0.05$).
Next, we carried out a Friedman chi-square test to compare models from the different paradigms. 
We chose RF and MHA from the previous analysis as representatives of \SNN and \TNN, respectively. The test reveals significant differences among the models. A post-hoc Conover test with a Bonferroni-adjusted $\alpha$ level confirmed that: 
\begin{itemize}
\item MHA significantly outperforms RF across all metrics and across all datasets.
\item \ReNN (GCN) significantly outperforms RF and LSTM in ACC@1, but not in ACC@Last on ELEA.
\item \TReNN significantly outperforms \ReNN (GCN) in both ACC@1 and ACC@Last (PAVIS), and also significantly outperforms \TNN (MHA) in ACC@Last (ELEA).
\end{itemize}

\emph{These results support our hypothesis that jointly modeling temporal and relational patterns yields a better generalization across both datasets. MHA significantly outperforms LSTM, likely due to its greater capacity to model long-range dependencies, particularly relevant here given that sequences often span over 20 time steps, while LSTMs are prone to information bottlenecks. This may suggest that identifying the least emergent leader depends more on temporal patterns (e.g., silence, speech activity) than on modeling relational information.} \\

\begin{table*}[t]
    \centering{ 
    \resizebox{\textwidth}{!}{
        \begin{tabular}{l|l|c|c|c|c|c|c|}
        \toprule
        & & \multicolumn{3}{c|}{PAVIS} & \multicolumn{3}{c|}{ELEA} \\
      \textbf{Paradigm} & \textbf{Model}   & \textbf{Dominance} & \textbf{Friendliness} & \textbf{Task orient.} & \textbf{Dominance} & \textbf{Friendliness} & \textbf{Task orient.}  \\
        \midrule
        \multirow{2}{*}{\centering \SNN} 
            & \col{RF} &  \col{\mstd{0.661}{0.008}} & \col{\mstd{0.605}{0.009}} & \col{\mstd{1.023}{0.007}} & \col{\mstd{0.729}{0.010}} & \col{\mstd{0.738}{0.009}} & \col{\mstd{0.958}{0.008}}  \\
             & FFNN  & \mstd{0.738}{0.009} & \mstd{0.652}{0.011} & \mstd{1.245}{0.011} & \mstd{0.758}{0.013} & \mstd{0.782}{0.0010} & \mstd{0.933}{0.015}\\
            \midrule
                    \multirow{2}{*}{\centering \TNN} 

             & \col{LSTM} & \col{\mstd{0.589}{0.006}} & \col{\mstd{0.592}{0.005}} & \col{\mstd{0.767}{0.007}} & \col{\mstd{0.685}{0.009}} & \col{\mstd{0.715}{0.011}} & \col{\mstd{0.891}{0.009}}\\
            & MHA & \mstd{0.538}{0.005}& \mstd{0.548}{0.008}& \mstd{0.738}{0.005}& \mstd{0.678}{0.004}& \mstd{0.643}{0.009}& \mstd{0.843}{0.008}\\
            \midrule
            
            \multirow{1}{*}{\centering \ReNN} 

            & \col{GCN} & \col{\mstd{0.513}{0.004}}  & \col{\mstd{0.503}{0.005}} &\col{\mstd{0.701}{0.004}}
            &\col{\mstd{0.638}{0.008}} & \col{\mstd{0.652}{0.010}} & \col{\mstd{0.839}{0.009}}\\  
        \midrule
            \TReNN & \col{GCN+MHA} & \col{\mstd{\textbf{0.488}}{0.005}} & \col{\mstd{\textbf{0.485}}{0.006}} & \col{\mstd{\textbf{0.675}}{0.007}}
             &\col{\mstd{\textbf{0.575}}{0.005}} & \col{\mstd{\textbf{0.582}}{0.006}} & \col{\mstd{\textbf{0.788}}{0.008}}\\
        \midrule      
        \bottomrule
    \end{tabular}
    }
    }
    \caption{Average MSE of LS (i.e., Dominance, Friendliness, Task orientation) over 10 different random seeds. The significantly best-performing method is boldfaced. Different models are grouped according to their main modeling paradigm (i.e., \SNN, \TNN, \ReNN, \TReNN)}.
    \label{tab:symlog_mse}
\end{table*}

\begin{table*}
    \centering{ 
    \resizebox{\textwidth}{!}{
        \begin{tabular}{l|c|c|c|c|c|c|c|c|c|}&
        \multicolumn{8}{c|}{PAVIS} \\
        Paradigm & Model   & A & BB & CC & MPM
&MT & SMM & TL & TO  \\
        \toprule        
        \multirow{2}{*}{\centering \SNN} 

 & RF &  \mstd{1.230}{0.011} & \mstd{1.207}{0.009} & \mstd{1.202}{0.010} & \mstd{1.229}{0.009} & \mstd{1.225}{0.008} & \mstd{1.221}{0.009} & \mstd{1.235}{0.008} & \mstd{1.229}{0.009}\\
 & \col{FFNN} &\col{\mstd{0.976}{0.010}} & \col{\mstd{1.087}{0.011}} & \col{\mstd{1.001}{0.009}} & \col{\mstd{0.905}{0.008}} & \col {\mstd{1.028}{0.011}} & \col{\mstd{1.122}{0.013}} & \col{\mstd{1.084}{0.011}} & \col{\mstd{1.109}{0.009}}\\
        \midrule
                \multirow{2}{*}{\centering \TNN} 

            & LSTM &  \mstd{0.765}{0.009} & \mstd{0.763}{0.007} & \mstd{0.772}{0.008} & \mstd{0.758}{0.009} & \mstd{0.854}{0.012} & \mstd{0.844}{0.009} & \mstd{0.856}{0.011} & \mstd{0.824}{0.013} \\
             & \col{MHA} &  \col{\mstd{0.795}{0.008}} & \col{\mstd{0.778}{0.009}} & \col{\mstd{0.772}{0.008}} & \col{\mstd{0.787}{0.009}} & \col{\mstd{0.892}{0.010}} & \col{\mstd{0.816}{0.011}} & \col{\mstd{0.844}{0.010}} & \col{\mstd{0.887}{0.012}} \\        \midrule
            \ReNN & GCN &  \mstd{0.731}{0.008} & \mstd{\textbf{0.726}}{0.009} & \mstd{\textbf{0.731}}{0.008} & \mstd{0.763}{0.007} & \mstd{0.909}{0.007} & \mstd{0.855}{0.006} & \mstd{0.834}{0.008} & \mstd{0.875}{0.005}\\   
        \midrule
            \TReNN & \col{GCN + MHA} &  \col{\mstd{\textbf{0.715}}{0.006}} & \col{\mstd{0.740}{0.007}} & \col{\mstd{0.752}{0.008}} & \col{\mstd{\textbf{0.729}}{0.005}} & \col{\mstd{\textbf{0.824}}{0.008}} & \col{\mstd{\textbf{0.798}}{0.009}} & \col{\mstd{\textbf{0.825}}{0.007}} & \col{\mstd{\textbf{0.768}}{0.007}}\\
        \midrule      
        \bottomrule
    \end{tabular}
    }
    }
    \caption{Average MSE of the TW components over 10 different random seeds. The significantly best-performing method is boldfaced. Different models are grouped according to their main modeling paradigm (i.e., \SNN,\TNN, \ReNN, \TReNN).}
    \label{tab:teamwork_light}
\end{table*}

\paragraph{\textit{Leadership Style}}
We formulated the LS prediction task of each team member as a multi-output regression problem with predictors and predicted variables, the paralinguistic features, and the scores extracted by SYMLOG, respectively.
Table~\ref{tab:symlog_mse} reports the MSE for each dimension. 
RF is implemented using 60 trees. The FFNN is designed as a three-layer architecture with the embedding size set to 16, the weight decay of 0.0125 to mitigate overfitting, and the learning rate set to 0.006 in the case of PAVIS, while in the case of ELEA, the learning rate is set to 0.0025 and the weight decay to 0.008. The LSTM model is configured as a four-layer architecture with embedding size set to 32, a weight decay of 0.02, and a learning rate of 0.0013 in the case of PAVIS, while the learning rate of 0.0009 for ELEA. The MHA model employs four attention heads within a four-layer architecture with the embedding size set to 16, with a weight decay of 0.0175 and a learning rate of 0.0015 for both datasets. The GCN is implemented as a three-layer architecture with embedding size set to 16, a weight decay of 0.0075, and the learning rate of 0.0075 in the case of PAVIS, while the weight decay is 0.01 and the learning rate 0.006 in the case of ELEA. Finally, \TReNN combines a two-layer GCN and a two-layer MHA with four attention heads, with embedding size set to 64, using a learning rate of 0.003 and a weight decay of 0.005 in the case of PAVIS, while learning rate 0.001 and weight decay 0.007 for ELEA.
The reported performance scores are averaged over 10 different random seeds. As for EL, we compare the performances of the different models among them. Data deviation from normality suggested the adoption of non-parametric tests. 
First, we assessed the differences within SNN and TNN paradigms using Hotelling’s $T^2$ test. The test shows that RF significantly outperforms FFNN and MHA significantly outperforms LSTM under both datasets, ELEA and PAVIS.
Next, we compared the top-performing models for SNN and TNN paradigms with the other models using a PERMANOVA test with 1000 permutations. This analysis revealed significant differences among the models. A post hoc Pairwise PERMANOVA test with a Bonferroni-adjusted $\alpha$ level shows that:
\begin{itemize}
\item MHA and \ReNN (GCN) both significantly outperform RF in both ELEA and PAVIS,
\item \ReNN (GCN) significantly outperforms MHA in both ELEA and PAVIS,
\item \TReNN (GCN+MHA) significantly outperforms \ReNN (GCN)  in both ELEA and PAVIS.
\end{itemize}

\emph{The findings support our hypothesis that both temporal and relational information are crucial for accurately modeling LS. Moreover, they indicate that relational information may play a comparatively stronger role in shaping leadership-related constructs. } \\

\begin{figure*}
    \centering
    \includegraphics[width=\textwidth]{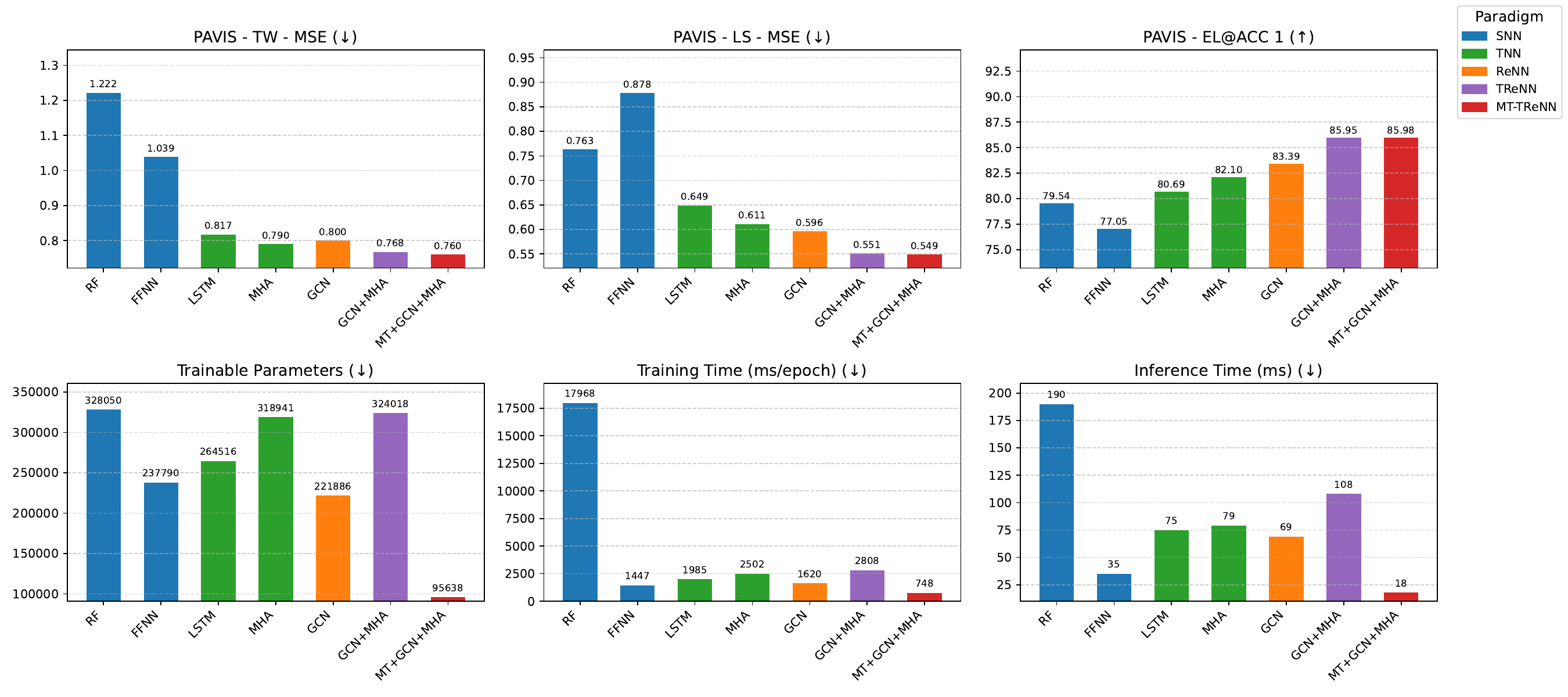}
    \caption{ Models'performance (PAVIS). In the first row, the MSE values of TW and of LS, and EL@ACC 1. In the second row, the trainable parameters required to achieve the performances, the training time (ms) for a single training epoch, and the inference time for a single team.}
    \label{fig:PAVIS_MTL}
\end{figure*}
\begin{figure*}
    \centering
    \includegraphics[width=\textwidth]{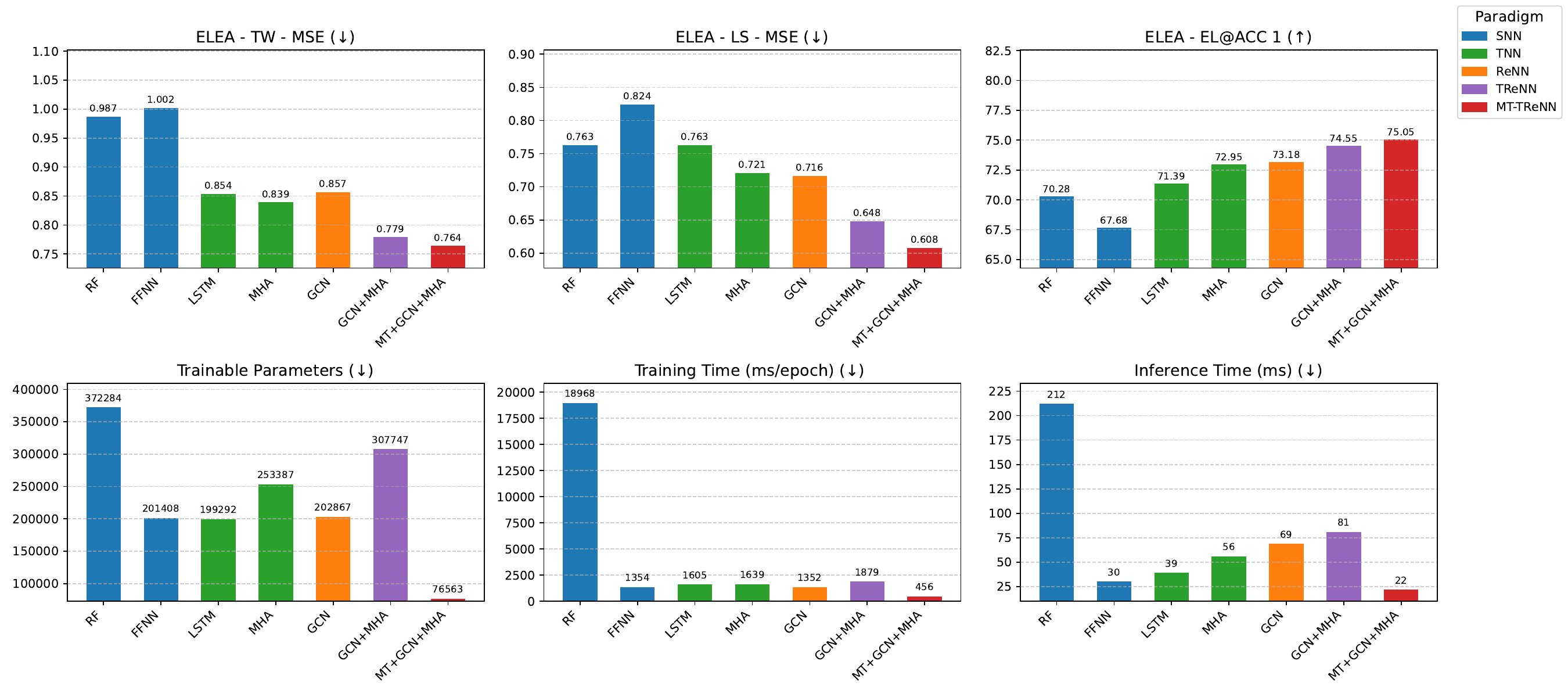}
    \caption{ Models'performance (ELEA). In the first row, the MSE values of TW and of LS, and EL@ACC1. In the second row, the trainable parameters required to achieve the performances, the training time (ms) for a single training epoch, and the inference time for a single team.}
    \label{fig:ELEA_MTL}
\end{figure*}

\paragraph{\textit{Teamwork}}

We formalized the prediction of the TW components as a set of regression tasks in which the predicted variables are the scores of each Teamwork component, computed as the average of the corresponding sub-scale of the BTF questionnaire. Table \ref{tab:teamwork_light} shows the values of MSE 
over 10 runs with different random seeds. 
RF is implemented using 75 trees. The FFNN is designed as a two-layer architecture with 16 as the embedding size, with a weight decay of 0.0075 and a learning rate of 0.002 for PAVIS, and a weight decay of 0.01 with a learning rate of 0.001 for ELEA.
The LSTM model adopts a four-layer architecture with 32 as the embedding size, using a weight decay of 0.02 and a learning rate of 0.0075 for PAVIS, while for ELEA, a weight decay of 0.01 and a learning rate of 0.002 are applied.
The MHA model uses four attention heads within a three-layer architecture with 32 as embedding size, with a weight decay of 0.008 and a learning rate of 0.00075 for PAVIS, and a weight decay of 0.01 with a learning rate of 0.00125 for ELEA.
The GCN is implemented as a three-layer architecture with 32 as the embedding size, using a weight decay of 0.007 and a learning rate of 0.002 for PAVIS, and a weight decay of 0.0012 with a learning rate of 0.001 for ELEA.
Finally, \TReNN combines a two-layer GCN with a two-layer MHA (four attention heads)  with 64 as embedding size, using a learning rate of 0.001 and a weight decay of 0.0025 for PAVIS, and a learning rate of 0.0015 with a weight decay of 0.005 for ELEA.
A Hotelling’s $T^2$ test was used to verify the possible differences in the performance of the SNN and TNN models. 
The results show that FFNN outperforms RF, and LSTM outperforms MHA, for the SNN and TNN paradigms (p < 0.05) under both datasets.
Next, a PERMANOVA test with 1000 permutations was run among the best SNN and TNN models and the other ones. This analysis revealed significant differences among the models. Thus, a post hoc Pairwise PERMANOVA test with a Bonferroni-adjusted $\alpha$ correction shows that:
\begin{itemize}
\item LSTM and \ReNN (GCN)  both significantly outperform FFNN in both PAVIS and ELEA,
\item \ReNN (GCN)  significantly outperforms MHA in both PAVIS and ELEA,
\item \TReNN (GCN+MHA) significantly outperforms \ReNN (GCN)  in both PAVIS and ELEA.
\end{itemize}

\emph{These findings support our hypothesis that jointly modeling temporal and relational information is crucial for accurately modeling TW. 
\TReNN outperforms the other models in most teamwork dimensions, apart from the BB and CC components, for which it performs slightly worse than \ReNN. MHA significantly outperforms LSTM, likely due to its better capacity to model long-range dependencies, which is fundamental in such a case where we have a sequence of 20 time steps, while LSTM is prone to lose information in longer sequences due to information bottlenecks. Moreover, they indicate that relational information (GCN) may play a comparatively stronger role than temporal information (MHA) in shaping TW components, indeed, GCN significantly outperforms MHA.} \\

\textbf{Q2: Does joint learning of Multiple Tasks increase efficiency in team modeling?} To address Q2, we compared the performance of the \MTTReNN architecture with single-task alternatives.

Fig.~\ref{fig:PAVIS_MTL} and Fig.~\ref{fig:ELEA_MTL} report the results on the PAVIS and the ELEA datasets, respectively.
The hyperparameter tuning of RF, FFNN, LSTM, MHA, GCN, and GCN+MHA is unaltered with respect to the previous, while \MTTReNN is defined as a GCN+MHA with 13 multi-task heads as a decoder, according to the design proposed in Sec.~\ref{sec:method}. The learning rate is set to 0.00025, and the importance of each task is automatically learned as described in Sec.~\ref{sec:decoder}, the weight decay is set to 0.001, the GCN is composed of two layers, the MHA is a two-layer architecture with four heads, and the decoder is realized as a two-layer FFNN. The dropout is introduced in both the encoder and the FFNN.
The top row compares predictive performance: the left plot shows MSE across TW components, the center plot reports average LS MSE (Dominance, Friendliness, Task Orientation), and the rightmost plot presents EL accuracy (ACC@1). \MTTReNN and \TReNN consistently and significantly outperform all other architectures, while the performance improvement of \MTTReNN over \TReNN is not statistically significant according to Hotelling's $T^{2}$ test. The effectiveness of \MTTReNN is apparent when considering computational costs and runtimes see the panel in Fig.~\ref{fig:PAVIS_MTL} \and Fig.~\ref{fig:ELEA_MTL}. \MTTReNN achieves, indeed, the lowest number of trainable parameters (left panel), the fastest training per epoch (middle panel), and the shortest inference time (right panel) among all architectures. These benefits arise because, unlike task-specific models that require multiple forward passes over independently processed samples, \MTTReNN produces all predictions in a single forward pass over a unified team representation. When compared to \TReNN, the only other architecture achieving comparable predictive performance, \MTTReNN has 70\% fewer parameters, 75\% reduction in training time, and 85\% reduction in inference time. 

\emph{The experimental results show that joint modeling multiple team constructs provides substantial advantages in terms of computational costs while retaining the predictive performance of the single task tempo-relational alternative. These results support the existence of a shared, tempo-relational embedding space in which high-level team dynamics co-exist, providing a promising direction for future work on team representation learning.}

\begin{figure*}
    \centering
    \begin{subfigure}{0.9\textwidth}
        \centering
        \includegraphics[width=\textwidth]{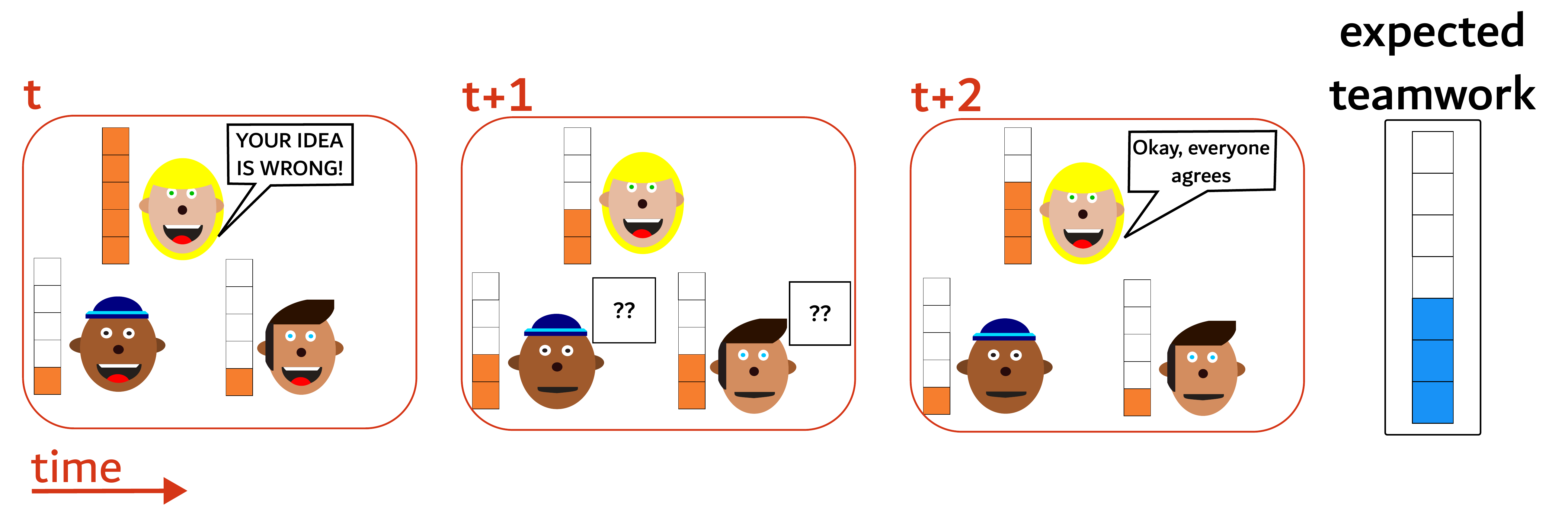}
        \caption{Example of Bad Teamwork}
        \label{fig:bad_teamwork}
    \end{subfigure}
    \begin{subfigure}{0.9\textwidth}
        \centering
        \includegraphics[width=\textwidth]{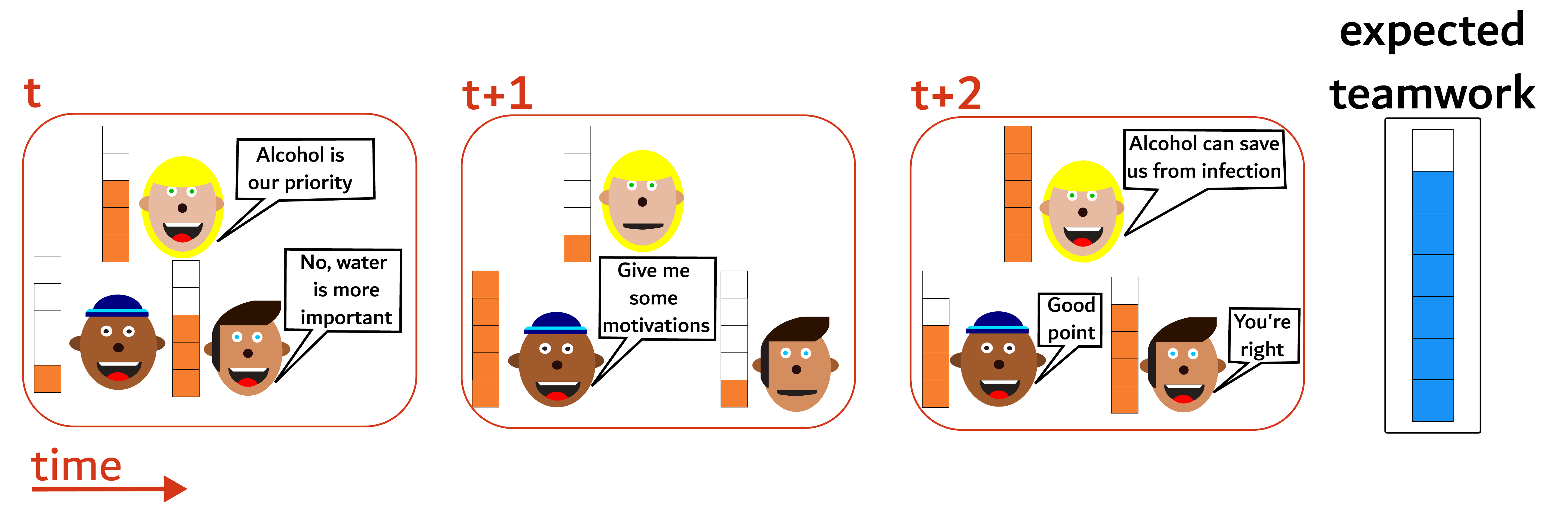}
        \caption{Example of Good Teamwork}
        \label{fig:good_teamwork}
    \end{subfigure}
    \caption{Examples of factual explanations for poor ((a)) and good ((b)) teamwork: orange bars indicate individual attribution scores, while the blue bar shows the expected teamwork quality.}
\label{fig:factual_explanation}
\end{figure*}

\paragraph{\bf Q3: Does \MTTReNN provide insights about team member contribution to team dynamics?}
\label{factual_explanations}

Understanding why a team achieves its performance, and how team dynamics could be improved, is crucial in high-stakes collaborative settings. These challenges can be addressed, respectively, through two complementary forms of explainability: factual and counterfactual. In the case of \SNN, explanations identify features or samples, in the case of \TNN, also temporal steps, while in graphs, explanations can target nodes and edges, in graph-based temporal models, explanations can naturally target nodes (team members), edges (interactions), and time steps. This structural richness makes \TReNN and \MTTReNN excellent candidates for producing two types of explanations. The first, factual explanations, provide a retrospective understanding of the factors driving model predictions. The second,  counterfactual explanations, identify actionable modifications to the interaction dynamics that could improve team performance. In our framework, both forms of explanations are grounded in the predicted teamwork effectiveness, computed as the average teamwork score across team members and teamwork components:}
\begin{equation}\label{expected_tw}
\mathbb{E}[\text{TW}] = \frac{1}{|G|}\sum_{v \in G}\frac{1}{m}\sum_{j=1}^{m} f_{trenn}(T,v)_{j}
\end{equation}
This formulation enables MT-TReNN to move beyond post-hoc interpretation towards decision-support in collaborative environments by grounding both explanation types in a unified quantitative measure of team effectiveness.

Factual explanations primarily support diagnostic understanding, highlighting which team members, interactions, and temporal segments most influenced the predicted teamwork score. Counterfactual explanations, instead, support actionable intervention, identifying minimal modifications to the interaction topology that are expected to improve teamwork. Taken together, these two forms of explanations allow \TReNN and \MTTReNN to move beyond prediction, providing interpretable and practically useful insights into team dynamics.
Note that this section does not aim to provide a quantitative evaluation of explanation quality, as such an assessment would require either ground-truth explanations or domain experts capable of judging candidate actions. Instead, our goal is to highlight the potential of the explainability component to identify meaningful patterns that can inform and improve team behavior.


\begin{figure*}
        \centering
        \includegraphics[width=.9\textwidth]{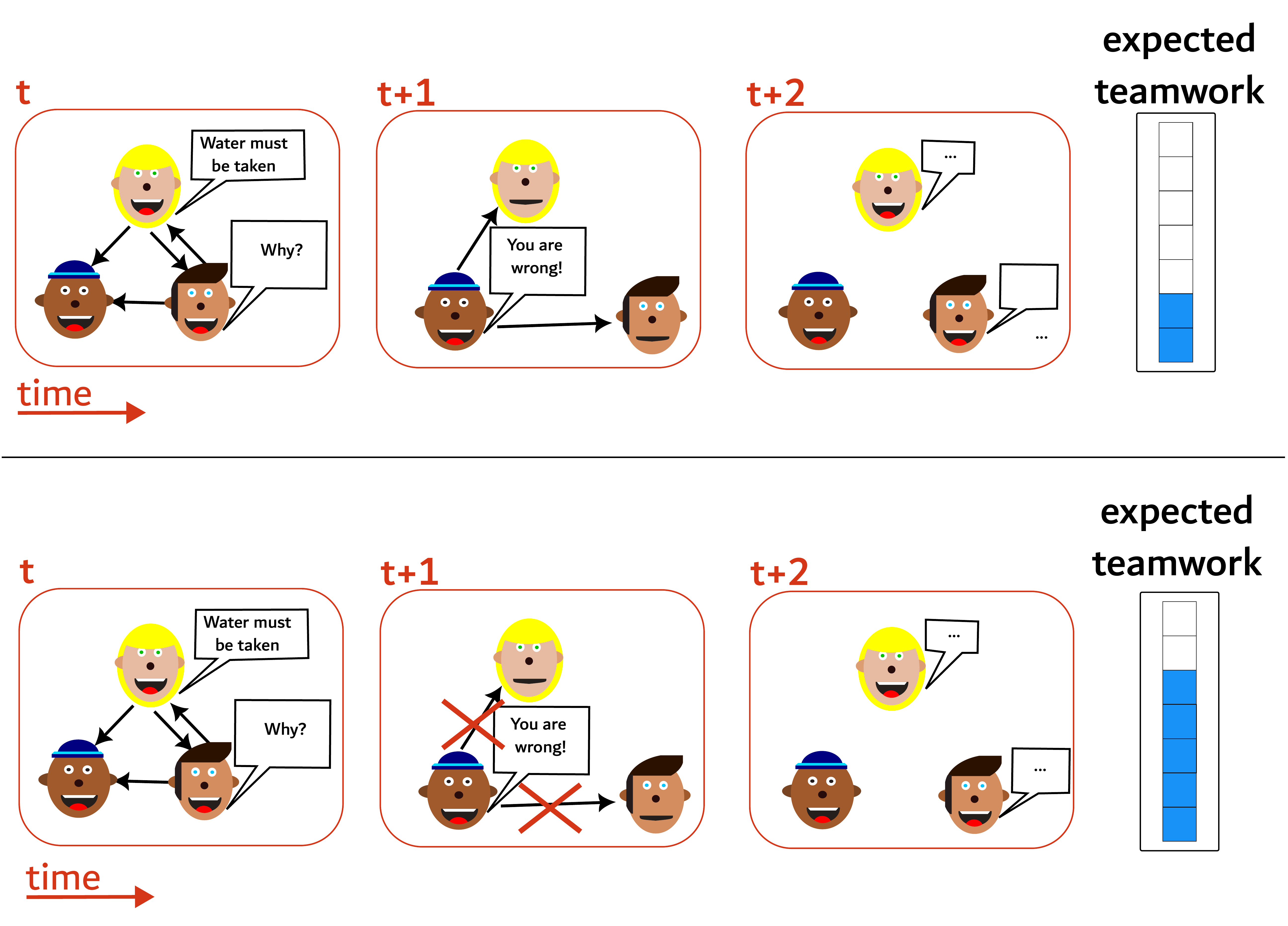}
        \caption{A counterfactual example: In the top panel, a team interaction dynamic that results in poor teamwork is depicted. In the bottom panel, the counterfactual explanation suggested by CoDy is reported.} 
        \label{fig:counterfactual_explanation}
\end{figure*}

\paragraph{\textit{Factual explanations}} 
As illustrated in Fig.~\ref{fig:complete_pipeline}, in an Operating Room scenario, an analyst may post-hoc inquire about leadership or mutual trust among team members. To support this type of analysis, we rely on factual post-hoc explanations that highlight which interactions and team members most influence the model’s predictions over time.
In this study, we employed an instance-based and gradient-based method, Saliency Map~\cite{simonyan2013deep}. Intuitively, Saliency Maps 
reflect how influential each input feature is for the model's output, which in our setting is the expected TW (Eq.~\ref{expected_tw}). The influence of a team member is computed by averaging its feature-level saliency scores. The rich topological structure of \TReNN and \MTTReNN enables insights from both temporal and relational perspectives, capturing how time and interactions influence individual team member-level predictions. 

Fig.~\ref{fig:factual_explanation} shows two anecdotal examples from the data: in each of them, team members are placed beside their attribution map score, which is discretized in five importance levels (orange bins), while the TW  
is shown on a seven-bin scale (blue bins).
In the top panel of the figure, we depict a scenario of bad teamwork. The team leader (the blonde girl) aggressively addresses the other team members (timestep \textit{t}), who refrain from responding, likely because of her aggressive tone (\textit{t+1}), so that she can unilaterally end the argument in her favour (\textit{t+2}). This situation is effectively captured by the attribution maps: the team leader has a maximal attribution score at the beginning and a high score at the end, while the other team members always have low scores, resulting in a low teamwork attribution score (blue bar). Conversely, the bottom panel illustrates a high-teamwork scenario. At timestep \textit{t}, two team members (i.e., the blonde girl and the black-haired guy) discuss which object is more important to bring with them to solve the task. At timestep \textit{t+1}, another team member (i.e., the one wearing a hat) asks for clarification, fostering a collaborative tone. At the timestep \textit{t+2}, the blonde girl offers a clear and well-reasoned explanation, prompting the agreement of the other two members. Individual attribution scores highlight the active participation of each team member, especially the request for clarification of the boy with the hat and the thoughtful response by the girl, resulting in a high average teamwork score.

\emph{Factual explanations generated via saliency maps enable an interpretable understanding of which team members and interactions most strongly influence the model’s predictions over time. This exemplifies how \TReNN can capture meaningful temporal and relational patterns that align with domain knowledge, making the model’s reasoning process transparent and trustworthy.}

\paragraph{\textit{Counterfactual explanations}}
As illustrated in Fig.~\ref{fig:complete_pipeline}, in an Operating Room scenario, team interactions can exhibit inadequate behaviors that negatively affect teamwork. Such situations can be addressed through targeted suggestions provided by the SIA, either online or offline, with the goal of directly influencing team dynamics. To this end, we rely on counterfactual explanations that identify which interactions should be modified or avoided to promote more effective collaboration, thereby fostering awareness of detrimental behaviors during operations.
In our setting, we aim to detect critical interactions that can potentially lead to enhanced TW. 
This task can be achieved via greedy approaches acting on the topology of a dynamic team, like CoDy~\cite{qu2024greedy}. At each iteration, CoDy performs the following step: i) select a node based on a selection score balancing exploration and exploitation; (ii) simulate its prediction outcome; (iii) expand the node through its child nodes; (iv) backpropagate updated scores to guide the search toward minimal yet effective counterfactuals. In our setting, this approach detects those interactions that can increase TW the most.

Fig.~\ref{fig:counterfactual_explanation} depicts an example in which, at timestep \textit{t}, a team member suggests an action (bringing water in the Winter Survival task) and another team member asks for an explanation. At the next timestep, the third team member forcefully asserts his opposing view, discouraging the others from speaking and ultimately undermining teamwork. The counterfactual explanation suggests the removal of the intervention of the third team member as the best modification to improve TW.

\emph{Counterfactual explanations provide actionable insights into crucial relational dynamics whose modification or removal could enhance team performance, highlighting the practical value of \TReNN beyond pure prediction.}

\subsection{Limitations and future work}

Despite the promising results, this work presents several limitations that should be acknowledged.
First, the experimental evaluation is conducted on the limited number of datasets commonly adopted in the literature, which may constrain the generalizability of the findings.
Team interactions are inherently domain-dependent, and the datasets considered include a restricted number of teams, primarily observed in controlled settings. Future work will focus on validating \TReNN and \MTTReNN in larger-scale and in-the-wild team scenarios, as such data become publicly available.
Second, the temporal graph extractor relies exclusively on paralinguistic features, which provide only a partial view of team behavior and limit the expressiveness of the resulting social embeddings. Incorporating additional modalities, such as lexical or visual cues, represents a natural extension of the proposed framework.
Finally, although \TReNN and \MTTReNN enable factual and counterfactual explanations, the lack of ground-truth explanations limits their evaluation to qualitative analyses. Future work will explore human-centered evaluation protocols to assess the usefulness and trustworthiness of the generated explanations.
Overall, addressing these limitations will further strengthen the applicability, robustness, and interpretability of tempo-relational models for team behavior analysis.

\section{Conclusion}
\label{sec:conclusion}
We presented \TReNN, a novel tempo-relational architecture for team modeling that jointly captures the relational and temporal structure of team dynamics. \TReNN leverages an automatic temporal graph extractor to convert team behavioral data into temporal graphs, a tempo-relational encoder to generate an embedding from temporal graphs, and a decoder to predict a team construct from the generated embedding. \TReNN outperforms alternative approaches that rely solely on temporal or relational information on two state-of-the-art datasets on EL, LS, and TW prediction.
On top of \TReNN, we presented its multi-task extension, \MTTReNN, that simultaneously can predict multiple team constructs by replacing the traditional decoder with several multi-task heads. 

\MTTReNN shows predictive performance comparable to \TReNN while drastically reducing space and time complexity.
Beyond predictive performance, experiments show that these novel architectures can generate interpretable insights as factual explanations and actionable suggestions to enhance team performance as counterfactual explanations. While the lack of ground truth explanations prevents an extensive evaluation of the quality of the generated explanations, their actionability suggests the potential for on-field evaluations involving real teams and decision-making support tasks. Finally, the current experimental setting shows promising results using paralinguistic features only. Further improvements can be expected by incorporating more complex multimodal features.

\paragraph{\bf Author contribution}
\label{author_contribution} V.M.D.L., G.V., and A.P. contributed to the design and validation of the experimental setup, and writing the manuscript. V.M.D.L. carried out the technical implementation, while G.V. and A.P. reviewed the manuscript.

\paragraph{\bf Funding}
Funded by the European Union. Views and opinions expressed are however those of the author(s) only and do not necessarily reflect those of the European Union or the European Health and Digital Executive Agency (HaDEA). Neither the European Union nor the granting authority can be held responsible for them. Grant Agreement no. 101120763 - TANGO. 
AP and VMDL acknowledge the support of the MUR PNRR project FAIR - Future AI Research (PE00000013) funded by the NextGenerationEU.

\paragraph{\bf Data Availability} The datasets, PAVIS and ELEA, used and analyzed during the current study can be obtained by filling the request form required respectively by Istituto Italiano di Technologia Pattern Analysis and Computer Vision, and IDIAP Research Institute.

\section*{Acknowledgments}
We acknowledge Raffaella Sabrina Fellone and Giuseppe De Luca for supporting the annotation process. (Portions of)
the research in this paper used the ELEA Dataset made available by the Idiap Research Institute, Martigny,
Switzerland.

\paragraph{\bf Competing Interests} The authors declare no competing interests.

\bibliographystyle{ACM-Reference-Format}
\bibliography{sample-base}

\end{document}